\documentclass{article}

\usepackage{graphicx}
\usepackage{wrapfig}

\usepackage{tabulary}
\usepackage{multirow}
\usepackage[margin=1in]{geometry}
\usepackage{titlesec}
\usepackage[labelfont=bf,textfont=it,font=footnotesize]{caption}
\usepackage{parskip}
\usepackage{mathpazo}
\usepackage[super,comma,sort&compress,numbers]{natbib}
\usepackage{booktabs}
\usepackage{listings}
\usepackage{float}
\usepackage{subcaption}

\usepackage{amssymb}
\usepackage{amsthm}
\usepackage{amsmath}
\usepackage{nicefrac}        % compact symbols for 1/2, etc.
\usepackage{siunitx}

\graphicspath{{Figures/}}
\DeclareGraphicsExtensions{.pdf,.png,.jpg}

\titlespacing\section{0pt}{6pt plus 1pt minus 0pt}{3pt plus 1pt minus 0pt}
\titlespacing\subsection{0pt}{4pt plus 1pt minus 0pt}{3pt plus 1pt minus 0pt}
\titlespacing\subsubsection{0pt}{4pt plus 1pt minus 1pt}{3pt plus 1pt minus 1pt}
\titleformat{\section}{\large\bfseries\sffamily}{\thesection}{1em}{}
\titleformat{\subsection}{\normalsize\bfseries\sffamily}{\thesubsection}{1em}{}
\titleformat{\subsubsection}{\normalsize\bfseries\sffamily}{\thesubsubsection}{1em}{}

\usepackage[ruled]{algorithm2e}

\usepackage{flushend}

\usepackage{tabulary}
\usepackage{multirow}

\usepackage{array}
\usepackage{enumitem}
\usepackage{todonotes}
\usepackage{url}
\usepackage[T1]{fontenc}
\usepackage{hyperref}
\hypersetup{
    colorlinks,
    linkcolor={magenta},
    citecolor={blue},
    urlcolor={blue!80!black},
    breaklinks=true,
	plainpages=true
}
\newcommand{\cref}[2]{\hyperref[#2]{#1~\ref*{#2}}}
\newcommand{\colref}[2]{\hyperref[#2]{#1~\ref*{#2}}}
\newcommand{\algref}[1]{\colref{Algorithm}{#1}}
\newcommand{\eqnref}[1]{\colref{Equation}{#1}}
\newcommand{\figref}[1]{\colref{Figure}{#1}}
\newcommand{\subfigref}[2]{Figure~\ref{#1}#2}
\newcommand{\secref}[1]{\colref{Section}{#1}}
\newcommand{\tabref}[1]{\colref{Table}{#1}}
\newcommand{\coloredref}[2]{\hyperref[#2]{#1~\ref*{#2}}}
\newcommand{\coloredsubref}[3]{\hyperref[#2]{#1~\ref*{#2}{#3}}}
\newcommand{\comment}[1]{}

\renewcommand\vec[1]{\ensuremath\mathbf{#1}}

\newcommand{\nd}{\textit{NURBS-Diff}}

\begin{document}

\begin{center}
{\usefont{OT1}{phv}{b}{}\selectfont\Large{NURBS-Diff: A Differentiable Programming Module for NURBS}}

{\usefont{OT1}{phv}{}{}\selectfont\normalsize
{Anjana Deva Prasad$^1$, Aditya Balu$^1$, Harshil Shah$^1$,
Soumik Sarkar$^1$,\\Chinmay Hegde$^2$, Adarsh Krishnamurthy$^1$*}}

{\usefont{OT1}{phv}{}{}\selectfont\normalsize
{$^1$ Iowa State University\\
$^2$ New York University\\
}}
\end{center}

%\red{Limit 150}
\section*{Abstract}

Boundary representations (B-reps) using Non-Uniform Rational B-splines (NURBS) are the de facto standard used in CAD, but their utility in deep learning-based approaches is not well researched. We propose a differentiable NURBS module to integrate NURBS representations of CAD models with deep learning methods. We mathematically define the derivatives of the NURBS curves or surfaces with respect to the input parameters (control points, weights, and the knot vector). These derivatives are used to define an approximate Jacobian used for performing the ``backward'' evaluation to train the deep learning models. We have implemented our NURBS module using GPU-accelerated algorithms and integrated it with PyTorch, a popular deep learning framework. We demonstrate the efficacy of our NURBS module in performing CAD operations such as curve or surface fitting and surface offsetting. Further, we show its utility in deep learning for unsupervised point cloud reconstruction and enforce analysis constraints. These examples show that our module performs better for certain deep learning frameworks and can be directly integrated with any deep-learning framework requiring NURBS.

\subsection*{Keywords}

Differentiable NURBS Layer $|$
NURBS $|$
Geometric Deep Learning $|$
Surface Modeling

% Comment out for final accepted paper submission
% Commented for better review
%\linenumbers
%%%%%%%%%%%%%%%%%%%%%%%%%%%%%%%%%%%%%%%%%%%%%%%%%%%%%%%%%%%%%%%%%%%%%
\section{Introduction}\label{Sec:Introduction}

In modern CAD systems, a solid model is represented using boundary representation (B-Rep), where the solid boundaries are defined using spline surfaces. Non-Uniform Rational B-splines (NURBS) are the standard representation used for defining the spline surfaces~\citep{10.5555/265261}. NURBS surfaces offer a high level of control and versatility; they can also compactly represent the surface geometry. NURBS surfaces can represent more complex shapes than B\`ezier or B-splines due to the non-uniformity of the knot vectors and the non-linear transformation due to the weights assigned to the control points. In addition, the NURBS definition allows for local control via the knots and the control points and global control via the weights. On the other hand, deep learning for 3D Euclidean geometry is emerging as a critical and well-explored research area in engineering. This area includes fundamental computer vision works such as 3D shape reconstructions from point clouds or multi-view stereo and 3D semantic segmentation for shape understanding~\citep{mescheder2019occupancy,park2019deepsdf,niemeyer2020differentiable,deng2020cvxnet,chen2020bsp,peng2020convolutional,davies2020overfit,atzmon2020sal,groueix2018atlasnet}. While NURBS are the standard CAD representation in engineering, their utility in deep learning-based approaches is not well researched. Current 3D deep learning (DL) research focuses on converting standard CAD representations to geometric representations that are more amenable to machine learning (such as voxels, triangular meshes, etc.)~\citep{barill2018fast}. This conversion from the standard CAD geometries to other representations is often irreversible and is not trivial to incorporate with DL algorithms.

One of the main challenges in extending NURBS-based representation to deep learning is the differentiable programming of the NURBS evaluations. The main idea of differentiable programming is to define each operation in a neural network in a differentiable manner. Differentiable programming uses gradient-based approaches to optimize the neural network parameters, thereby obtaining the desired output through neural network operations. This differentiable programming paradigm allows an end-to-end programmable system that can be used to train deep neural networks. This paradigm has found use in a large variety of applications such as scientific computing~\citep{Innes2020ALGORITHMICD, Innes2019ADP, Schafer2020ADP}, image processing~\citep{li2018differentiableprog}, physics engines~\citep{Degrave2017ADP}, computational simulations~\citep{alnaes2015fenics}, and graphics~\citep{li2018differentiable, Chen2019LearningTP}.

\begin{figure}[t!]
    \centering
    \includegraphics[width=0.6\linewidth,trim={0.8in 1.0in 0.1in 1.0in},clip]{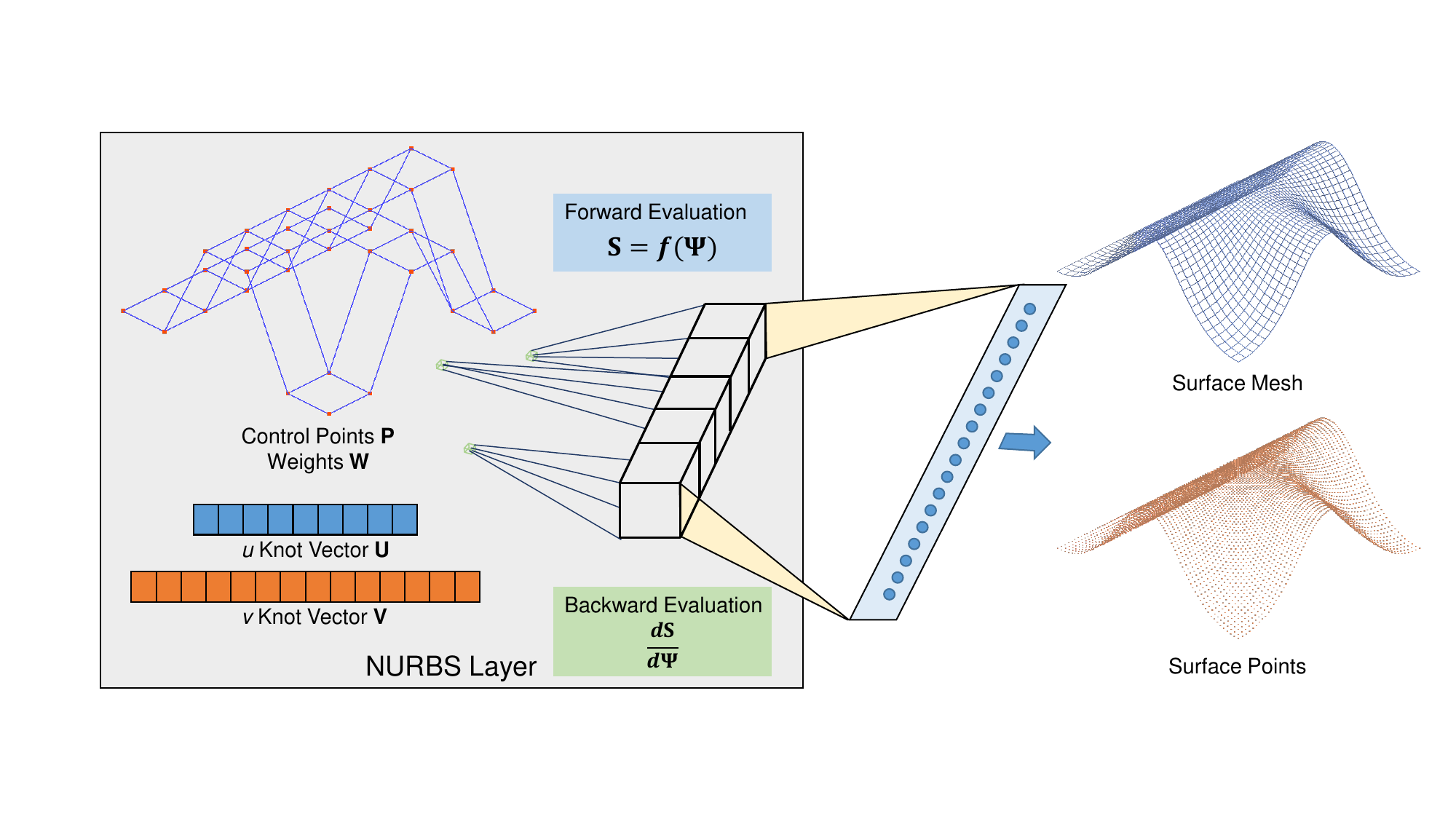}
    \caption{We propose a differentiable NURBS module that can be used for CAD geometric modeling using standard deep learning systems. The NURBS parameters are input to the module during the forward evaluation, and we evaluate the surface mesh. Once a loss is computed, and gradients for the surface mesh are obtained, we perform a backward evaluation to enable backpropagation of the losses to modify the input parameters.}
    \label{Fig:NURBSLayer}
\end{figure}

For differentiable programming of NURBS, we need to compute gradients of the NURBS surface points with respect to the parameters of the NURBS representation (see \figref{Fig:NURBSLayer}). However, since the NURBS surface points are a function of knots, control points, and weights, the partial derivative with respect to each input parameter needs to be computed, making the backward evaluation challenging. Further, due to the use of basis functions (which are piecewise-continuous polynomials), gradients might be discontinuous (or even zero) at the knots. The recursive definition of the basis functions adds an additional challenge to define derivatives with respect to the knot vectors. To alleviate these issues, we exploit the recent theoretical advances in the paradigm of differentiable programming. It has been theoretically proven that if a weak form of the Jacobian for the ``forward'' evaluation operator can be represented using a \textit{block-sparse} matrix, then this approximate Jacobian can be used to perform the ``backward'' evaluation of that operation~\cite{cuturi2019differentiable,chodifferentiable}. This approach has been used to define approximate derivatives for operations such as sorting~\citep{Blondel2020FastDS,cuturi2019differentiable}, loops and algorithmic conditions, and even derivatives for piecewise polynomial functions~\citep{baydin2018automatic}. We use a similar approach by defining a \textit{block-sparse} Jacobian for NURBS surface evaluation. Existing differentiable programming approaches for splines~\citep{Muller-2019} mainly focus on optimizing the control point locations. In this paper, we provide a complete module for integrating NURBS with differentiable frameworks that optimize not just the control points but also the knots for reparameterization.

Formally, using the differentiable programming approach explained above, we formulate a differentiable NURBS (``\nd{}'') module, which enables deep learning frameworks to integrate NURBS-based representation of B-Rep surfaces and perform CAD operations using them. The forward pass of our NURBS module uses the standard NURBS evaluation as explained in \secref{SubSec:Forward}. The backward pass uses the derivatives of the NURBS curve or surface with respect to the input parameters of the NURBS (\secref{SubSec:Backward}). The derivatives are used to define the Jacobian that is then used for backpropagation of the losses during training. After defining \nd{}, we validate our approach by performing traditional CAD operations such as curve fitting, surface fitting, and surface offsetting. Finally, we show the applicability of \nd{} in deep learning by using it as an additional decoder for point cloud reconstruction.

In this paper, we have developed a differentiable NURBS module that can be used in machine learning and CAD applications. The key contributions of this work are:
\begin{enumerate}[topsep=0pt,itemsep=0pt]
    \item A differentiable programming framework using NURBS representation where the losses can be back-propagated using the NURBS definitions. Specifically, we define the Jacobian based on the derivatives of the NURBS with respect to its input parameters.
    \item A GPU-accelerated implementation of our NURBS module in PyTorch for better integration with existing deep learning programming frameworks.
    \item A gradient descent-based optimization framework using \nd{} for performing CAD operations such as curve/surface fitting and surface offsetting. 
    \item The applicability of our proposed differentiable programming framework to extend the training process of unsupervised point cloud reconstruction of NURBS surfaces.
\end{enumerate}

The rest of the paper is arranged as follows. We outline some close related work in point cloud reconstruction, machine learning approaches in CAD, and differentiable programming in \secref{Sec:RelatedWork}. We provide the mathematical details of our differentiable NURBS module in \secref{Sec:NURBSLayer}. We show the application of \nd{} to CAD operations in \secref{Sec:CADApplications} and to unsupervised point cloud reconstruction in \secref{Sec:PCReconstructions}.

\section{Related Work}\label{Sec:RelatedWork}

The problem of extracting concise geometry representations from a spectrum of input data formats such as images, depth maps, and point clouds has been extensively studied over the last few decades. While methodologies that derive such representations are pervasive in 3D reconstruction literature today, our NURBS module focuses on filling the gap between the NURBS-based CAD representation and the other input formats used in machine learning. In this context, we broadly categorize the prior related work under differentiable programming and splines in deep learning. 

\subsection{Differentiable Programming}
NURBS surfaces are obtained as a tensor product of two-piecewise polynomial B-spline curves. To conceptualize end-to-end trainable deep learning systems that can fit NURBS surfaces to various input geometries, we require a framework that can backpropagate over such piecewise polynomial functions. Several recent works have been proposed that take advantage of the differentiable programming paradigm to approximate gradients for such functions. \citet{cuturi2019differentiable} and \citet{blondel2020fast} propose differentiable operators for sorting based tasks. Similarly, \citet{vlastelica2020differentiation} compute gradients for several optimization problems by constructing linear approximations to discrete-values functions. We model our module on prior work that incorporates structured priors as modules in the deep learning framework, similar to \citet{sheriffdeen2019accelerating}, \citet{Joshi2020InvNetEG}, and \citet{10.5555/3294771.3294868}. Beyond deep learning-based approaches, automatic differentiation for NURBS parametric coordinates for obtaining the surface derivatives for Adjoint-based sensitivity analysis has been performed by \citet{zhang2018cad}. \citet{ugolotti2021differentiated} performed a gradient-based aerodynamic shape optimization using a robust Machine Learning model, which is created to integrate the geometry generation and the mesh generation process using one single polynomial module for the volumetric mesh. \citet{mykhaskiv2018nurbs} and \citet{muller2019geometric} define a differentiated CAD kernel in OpenCASCADE for applying algorithmic differentiation to a complete CAD system for shape optimization and imposing constraints. These works encourage us to pursue a similar research direction in developing the \nd{} module. However, their approach for obtaining the gradients involves tedious operations such as performing singular value decomposition (SVD) on the control points to obtain the gradients. We also note that the mathematical definition of the derivatives for NURBS and its application to fitting has been explored previously~\citep{hoschek1988intrinsic,piegl1998computing}. However, due to discontinuities, a more stable and faster approach for computing the derivatives is needed.  

\subsection{Splines in Deep Learning}
Several deep learning frameworks use splines. \citet{minto2018deep} use NURBS surfaces fitted over the 3D geometry as an input representation for the object classification task of ModelNet10 and ModelNet40 datasets. \citet{erwinski2016neural} presented a neural-network-based contour error prediction method for NURBS paths. \citet{fey2018splinecnn} present a new convolution operator based on B-splines for irregular structured and geometric input, e.g., graphs or meshes. \citet{balestriero2018spline} build a theoretical link between deep networks and spline functions and build end-to-end deep learning systems using spline-based activation functions. \citet{balu2019deep} propose a NURBS-aware convolutional neural network that maintains the topological structure similar to a parametric NURBS surface evaluation grid. Very recently, \citet{sharma2020parsenet} performed point cloud reconstruction to predict a B-spline surface, which is later processed to obtain a complete CAD model with other primitives ``stitched'' together. In our work, we perform comparison between our approach and \citet{sharma2020parsenet} in \secref{Sec:PCReconstructions}.

\section{NURBS-Diff Module}\label{Sec:NURBSLayer}
Modern CAD systems make use of boundary-representation (B-Rep) for representing a solid geometry, $\Omega$, which is embedded in a 2D or 3D Euclidean space ($\Re^{2}$ or $\Re^{3}$). A B-Rep consists of a set of surfaces $d\Omega$ representing the boundary of the solid. Each surface $\mathcal{S} \subset d\Omega$ in a standard CAD system is represented using a Non-Uniform Rational B-spline (NURBS) surface. The NURBS representation is a compact representation that uses a set of control points, knot vectors, degrees, and weights to map a parametric space to span the entire surface $\mathcal{S}$ in the Euclidean space. In this work, we propose a differentiable NURBS module which could evaluate the surface $\mathcal{S}$ given the control points, knot vectors, degrees, and weights, usually obtained as an output from a deep learning system, $NN(\theta)$. This deep learning system is trained using a loss function $\mathcal{L}(\cdot,\cdot)$, computed between a target point cloud $\vec{P} \in  \{ \Re^{N x 2}$ or $\Re^{N x 3} \}$ (where N is the number of points) and a set of points $\vec{S}$ sampled (or evaluated) from the surface $\mathcal{S}$. During the training process, the gradient of the loss function with respect to the parameters of the deep learning model, $\nicefrac{\partial{\mathcal{L}}}{\partial{\theta}}$ is required for back propagation. It is usually straightforward to compute the derivative of the loss function $\nicefrac{\partial{\mathcal{L}}}{\partial{\mathcal{S}}}$ for some of the standard loss functions used, such as Chamfer distance, $L_2$ distance, etc., since they are differentiable. However, $\nicefrac{\partial{\mathcal{L}}}{\partial{\theta}}$ requires a mathematically consistent definition of $\nicefrac{\partial{\mathcal{S}}}{\partial{\Psi}}$, where $\Psi$ refers to the complete set of NURBS parameters (i.e. the set of control points $\mathbf{P}$, its corresponding weights $\mathbf{W}$, and the knot vectors $\mathbf{U}$ and $\mathbf{V}$) that define the surface. The gradient $\nicefrac{\partial{\mathcal{S}}}{\partial{\Psi}}$ is necessary because the deep learning system $NN(\theta)$ predicts this set of NURBS parameters $\Psi$ and computing $\nicefrac{\partial{\mathcal{L}}}{\partial{\theta}}$ requires computing $\nicefrac{\partial{\mathcal{S}}}{\partial{\Psi}}$, $\nicefrac{\partial{\mathcal{L}}}{\partial{\mathcal{S}}}$ and finally, $\nicefrac{\partial{\Psi}}{\partial{\theta}}$. Formally, this can be explained using the chain rule as:
\begin{equation}
       \frac{\partial{\mathcal{L}}}{\partial{\theta}} =  \frac{\partial{\mathcal{L}}}{\partial{\mathcal{S}}}\, \frac{\partial{\mathcal{S}}}{\partial{\Psi}}\, \frac{\partial{\Psi}}{\partial{\theta}}
       \label{eq:PartialDerivatives}
\end{equation}

The main challenge in this approach is computing $\nicefrac{\partial{\mathcal{S}}}{\partial{\Psi}}$. To this end, we propose a differentiable NURBS module implemented as a forward and backward machine learning module. While our module can handle both curve and surface point computations, we limit the discussions of our forward and backward algorithms to surfaces. As shown in the results section, the approach can be directly used for curves embedded in both 2D and 3D space by suitably adjusting the dimensions of the NURBS parameters.

\subsection{Forward Evaluation for NURBS Surface}\label{SubSec:Forward}
The NURBS surface $\mathcal{S}$ is sampled over a finite parametric space $(u,v)$ where $(u,v)\in ([0,1] \times [0,1])$, and this set of finite points $\vec{S}$ representing the surface is used for performing the loss computation and the backward gradient computation. This set $\vec{S}$ is computed as a function of NURBS parameters $\Psi = \{\mathbf{P},\mathbf{U}, \mathbf{V},\mathbf{W}\}$. Given the NURBS surface points are a function of the NURBS parameters in \eqnref{eq:ForwardEvaluation}, we compute the forward evaluation using the NURBS formulation:
\begin{equation}
    \vec{S} =\mathbf{f} \left( \vec{P}\,,\vec{U}\,,\vec{V}\,,\vec{W} \right)
    \label{eq:ForwardEvaluation}
\end{equation}

\subsubsection {NURBS Formulation}
Formally, a point in the NURBS surface parametrized using $(u,v)$ is defined as follows:
\begin{equation}
	\vec{S}(u,v)=\frac{\sum_{i=0}^n{\sum_{j=0}^m{N_i^p(u) N_j^q(v)w_{ij}\vec{P}_{ij}}}}{\sum_{i=0}^n{\sum_{j=0}^m{N_i^p(u) N_j^q(v)w_{ij}}}} , 
	\label{eq:NURBS}
\end{equation}

Here, the basis functions of NURBS, ($N_i, N_j$) are polynomials that are recursively computed using Cox-de Boor recursion formula in \eqnref{eq:NURBSBasis}, where $u$ is the parameter value, $N_i^p$ is the $i^{th}$ basis function of degree $p$.
\begin{equation}
	N_i^p(u)=\frac{u-u_i}{u_{i+p}-u_i}N_i^{p-1}(u)+\frac{u_{i+p+1}-u}{u_{i+p+1}-u_{i+1}}N_{i+1}^{p-1}(u)
	\label{eq:NURBSBasis}
\end{equation}
\begin{equation}
	N_i^0(u) = \left\{
	\begin{array}{l l}
		1  \quad \mbox{if $u_i\leq u\leq u_{i+1}$} \\
		0  \quad \mbox{otherwise}
	\end{array} \right.
	\label{eq:BSplineBasis2}
\end{equation}
Here, $u_i$ (also known as knots) refers to the elements of the knot vector $\vec{U}$ (similarly, $v_i \in \vec{V}$). The knot vector is a non-decreasing sequence of parametric coordinates, which divides the B-spline into non-uniform piecewise functions. The basis functions $N_i^p$ spans over the parametric domain based on the knot vector and degree as shown in \eqnref{eq:NURBSBasis} and \eqnref{eq:BSplineBasis2}. Note that the formulation explained in \eqnref{eq:NURBS} uses the vector notation, where $\vec{P}_{ij}$ is embedded in $\Re^3$. 

\subsubsection{Surface Point Evaluation}

The complete algorithm for forward evaluation of $\Vec{S}(u,v)$ as described in \citet{10.5555/265261} can be divided into three steps: 
\begin{enumerate}
\item Finding the knot span of $u \in [u_i,u_{i+1})$ and the knot span of $v \in [v_j,v_{j+1})$, where $u_i, u_{i+1} \in \mathbf{U}$ and $v_ j, v_{j+1} \in \mathbf{V}$. This is required for the efficient computation of only the non-zero basis functions.
\item Now, we compute the non-zero basis functions $N_i^p(u)$ and $N_j^q(v)$ using the knot span. The basis functions have specific mathematical properties that help us evaluate them efficiently. The partition of unity and the recursion formula ensure that the basis functions are non-zero only over a finite span of $p+1$ control points. Therefore, we only compute those $p+1$ non-zero basis functions instead of the entire $n$ basis function. Similarly in the $v$ direction we only compute $q+1$ basis functions instead of $m$.
\item We first compute the weighted control points $\vec{P}^w_{ij}$ for a given control point $\vec{P}_{ij}=\{\vec{P}_x, \vec{P}_y , \vec{P}_z\}$ and weight $w_{ij}$ as $\{\vec{P}_x w, \vec{P}_y w, \vec{P}_z w\}$ representing the surface after homogeneous transformation for ease of computation. Once the basis functions are computed we multiply the non-zero basis functions with the corresponding weighted control points, $\vec{P}^w_{ij}$. This result, $\vec{S'}$ is then used to compute $\vec{S}(u,v)$ as $\{ S'_{x}/S'_{w}, S'_{y}/S'_{w}, S'_{z}/S'_{w}\}$.
\end{enumerate}

\subsubsection{Implementation}
In a deep learning system, each module is considered an independent unit that performs the computation. During the forward pass, the module takes a batch of input and transforms them using the parameters of the module parameters. Further, to reduce the computations needed during the backward pass, we store extra information for computing the gradients during the forward computation. The \nd{} module takes as input the control points, weights, and knot vectors for a batch of NURBS surfaces. We define a parameter to control the number of points evaluated from the NURBS surface. We define a mesh grid of a uniformly spaced set of parametric coordinates $u_{grid}\times v_{grid}$. We perform a parallel evaluation of each surface point $S(u,v)$ in the $u_{grid}\times v_{grid}$ for all surfaces in the batch and store all the required information for the backward computation. The complete algorithm is shown in \algref{Alg:Forward}.

\begin{algorithm}[h!]
    \caption{Forward algorithm for multiple surfaces\label{Alg:Forward}}
    \SetKwInOut{Input}{Input}
    \SetKwInOut{Output}{Output}

    \Input{$\vec{U}$, $\vec{V}$, $\vec{P}$, $\vec{W}$, output resolution $n_{grid}$, $m_{grid}$}
    \Output{$\vec{S}$}
    
    \text{Initialize a meshgrid of parametric coordinates}
    \text{uniformly from $[0,1]$ using $n_{grid}\times m_{grid}$ : $u_{grid} \times v_{grid}$}\\
    \text{Initialize: $\vec{S} \rightarrow \vec{0}$}\\
    \For{$k = 1: surfaces$ in \textbf{\emph{parallel}}}
    {
    \For{$j=1:m_{grid}$ points in \textbf{\emph{parallel}}}
      {
        \For{$i=1:n_{grid}$ points in \textbf{\emph{parallel}}}
          {
            Compute $u_{span}$ and $v_{span}$ for the corresponding $u_i$ and $v_i$ using knot vectors $\vec{U_k}$ and $\vec{V_k}$ \\
            Compute basis functions $N_i$ and $N_j$ basis functions using $u_{span}$ and $v_{span}$ and knot vectors $\vec{U_k}$ and $\vec{V_k}$\\
            Compute surface point $\vec{S}(u_i,v_j)$ (in $x$, $y$, and $z$ directions).\\
            Store $u_{span}$, $v_{span}$, $N_i^p$, $N_j^q$, and $\vec{S}(u_i,v_j)$ for backward computation
          }
      }
    }
\end{algorithm}

Our implementation is robust and modular for different applications. For example, if an end-user desires to use this for a B-spline evaluation, they need to set the knot vectors to be uniform and weights $\vec{W}$ to be $1.0$. In this case, the forward evaluation can be simplified to $\vec{S}(u,v) = \vec{f}(\vec{P})$. Further, we can also pre-compute the knot spans and basis functions during the initialization of the NURBS-Diff module. During computation, we could use tensor comprehension that significantly increases the computational speed. We can also handle NUBS (Non-Uniform  B-splines), where the knot vectors are still non-uniform, but the weights $W$ are set to $1.0$. Note in the case of  B-splines $\Psi = \{\vec{P}\}$ (the output from the deep learning framework) and in the case of NUBS $\Psi = \{\vec{P}, \vec{U}, \vec{V}\}$.

\subsection{Backward Evaluation for NURBS Surface}\label{SubSec:Backward}
In a modular machine learning system, each computational module requires the gradient of a loss function with respect to the output tensor for the backward computation or the backpropagation. For our NURBS-Diff module this corresponds to $\nicefrac{\partial{\mathcal{L}}}{\partial{\mathcal{S}}}$ . As an output to the backward pass, we need to provide $\nicefrac{\partial{\mathcal{L}}}{\partial{\Psi}}$. While we represent $\mathcal{S}$ for the boundary surface, computationally, we only compute $\vec{S}$ (the set of surface points evaluated from $\mathcal{S}$). Therefore, we would be using the notation of ${\partial{\vec{S}}}$ instead of ${\partial{\mathcal{S}}}$ to represent the gradients with respect to the boundary surface. Here, we assume that with increasing the number of evaluated points, ${\partial{\vec{S}}}$ will asymptotically converge to ${\partial{\mathcal{S}}}$. Now, we explain the computation of $\nicefrac{\partial{\vec{S}}}{\partial{\Psi}}$ in order to compute $\nicefrac{\partial{\mathcal{L}}}{\partial{\Psi}}$ using the chain rule. To explain the implementation of the backward algorithm, we first explain the NURBS derivatives for a given surface point with respect to the different NURBS parameters.

\subsubsection {NURBS Derivatives}
We rewrite the NURBS formulation as follows:
\begin{equation}
\vec{S}(u,v) = \frac{\vec{NR}(u,v)}{w(u,v)}
\label{eq:NURBSsplit}
\end{equation}
where,  
\begin{equation*}
\vec{NR}(u,v) = \sum_{i=0}^n{ \sum_{j=0}^m{N_i^p(u) N_j^q(v)w_{ij}\vec{P}_{ij} } }
\end{equation*}

\begin{equation*}
{w}(u,v) = \sum_{i=0}^n\sum_{j=0}^m{N_i^p(u) N_j^q(v)w_{ij}}
\end{equation*}

For the forward evaluation of $\vec{S}(u,v) =\mathbf{f} \left( \vec{P}\,,\vec{U}\,,\vec{V}\,,\vec{W} \right)$, we can define four derivatives for a given surface evaluation point: $\vec{S}_{,u} :=  \nicefrac{\partial{\vec{S}(u,v)}}{\partial{u}}$, $\vec{S}_{,v} :=  \nicefrac{\partial{\vec{S}(u,v)}}{\partial{v}}$, $\vec{S}_{,\vec{P}} :=  \nicefrac{\partial{\vec{S}(u,v)}}{\partial{\vec{P}}}$, and $\vec{S}_{,\vec{W}} :=  \nicefrac{\partial{\vec{S}(u,v)}}{\partial{\vec{W}}}$. Note that $\vec{S}_{,\vec{P}}$ and $\vec{S}_{,\vec{W}}$ are represented as a vector of gradients $\{\vec{S}_{,P_{ij}} \forall P_{ij} \in \vec{P}\}$ and $\{\vec{S}_{w_{ij}} \forall w_{ij} \in \vec{W}\}$.

Now, we show the mathematical form of each of these four derivatives. The first two derivatives are traditionally known as the parametric surface derivatives, $\vec{S}_{,u}$ and $\vec{S}_{,v}$. Here, $N_{i,u}^p(u)$ refers to the derivative of basis functions with respect to $u$ and $v$, respectively. These are the standard parametric derivatives, and we do not repeat them here; they are provided in the Appendix for completeness. These derivatives are useful in the sense of differential geometry of NURBS for several CAD applications~\citep{Krishnamurthy-2009}. However, we do not use it in our module since many deep learning applications such as surface fitting are not dependent on the $(u,v)$ parametric coordinates. Also, note that $\vec{S}_{,u}$  and $\vec{S}_{,v}$ are not the same as $\vec{S}_{,\vec{U}}$ and $\vec{S}_{,\vec{V}}$. The formulation for $\vec{S}_{,\vec{U}}$ and $\vec{S}_{,\vec{V}}$ is provided later in this section.

Now, let us define $\vec{S}_{,p_{ij}}(u,v)$:
\begin{equation}
	\vec{S}_{,\vec{P}_{ij}}(u,v)= \frac{N_i^p(u) N_j^q(v)w_{ij}}{\sum_{k=0}^n\sum_{l=0}^m{N_k^p(u) N_l^q(v)w_{kl}}}
	\label{eq:NURBSuDerivative4}
\end{equation}
where $\vec{S}_{,\vec{P}_{ij}}(u,v)$ is the rational basis functions themselves. Computing $\vec{S}_{,w_{ij}}(u,v)$ is more involved with $w_{ij}$ terms in both the numerator and the denominator of the evaluation.
\begin{equation}
	\vec{S}_{,w_{ij}}(u,v)=\frac{\vec{NR}_{,w_{ij}}(u,v)w(u,v) - \vec{NR}(u,v)w_{,w_{ij}}(u,v)}{w(u,v)^2}
	\label{eq:NURBSvDerivative5}
\end{equation}
where, 
\begin{equation*}
	\vec{NR}_{,w_{ij}}(u,v)={N_{i}^p(u) N_j^q(v)\vec{P}_{ij}}
\end{equation*}
\begin{equation*}
    {w}_{,w_{ij}}(u,v) = {N_{i}^p(u) N_j^q(v)}
\end{equation*}

For the forward evaluation of $\vec{S}(u,v) =\mathbf{f} \left( \vec{P}\,,\vec{U}\,,\vec{V}\,,\vec{W} \right)$, we have defined $\vec{S}_{,\vec{P}}(u,v)$ and $\vec{S}_{,\vec{W}}(u,v)$ along with the derivatives $\vec{S}_{,u}(u,v)$ and $\vec{S}_{,v}(u,v)$. However, computing the $\vec{S}_{,\vec{U}}(u,v)$ and $\vec{S}_{,\vec{V}}(u,v)$ is not trivial. $\vec{S}_{,\vec{U}}(u,v)$ and $\vec{S}_{,\vec{V}}(u,v)$ refer to the $\nicefrac{\partial \vec{S}(u,v)}{\partial u_i}, \; s.t. u_i\in \vec{U}$ and $\nicefrac{\partial \vec{S}(u,v)}{\partial v_i}, \; s.t. v_i\in \vec{V}$. $\vec{U}$ and $\vec{V}$ influence the computation of the basis functions, and these derivatives are helpful for reparameterization of the surfaces by changing the knot vectors. However, due to the recursive computation of the basis functions, the derivatives for $\vec{U}$ and $\vec{V}$ are not defined. Therefore, we need a more rigorous approach for defining the differentiable programming of knot vectors.

First, lets decompose the derivative $\nicefrac{\partial \vec{S}(u,v)}{\partial u_i}, \; s.t. u_i\in \vec{U}$\footnote{We explain the formulation for $\vec{U}$; a similar formulation exists for $\vec{V}$.} into the derivative of $\nicefrac{\partial \vec{S}(u,v)}{\partial N^p_i(u)}$ and the partial derivative of $\nicefrac{\partial N^p_i(u)}{\partial u_i}$. The derivative $\nicefrac{\partial \vec{S}(u,v)}{\partial N^p_i(u)}$ can be easily computed from chain rule as shown here:

\begin{align}
 	& \frac{\partial \vec{S}(u,v)}{\partial N^p_i(u)}  = \frac{{\sum_{j=0}^m{ N_j^q(v)w_{ij}\vec{P}_{ij}}}}{\sum_{r=0}^n{\sum_{j=0}^m{N_r^p(u) N_j^q(v)w_{rj}}}} \nonumber \\ &   - \frac{\sum_{r=0}^n{\sum_{j=0}^m{N_r^p(u) N_j^q(v)w_{rj}\vec{P}_{rj}}}}{(\sum_{r=0}^n{\sum_{j=0}^m{N_r^p(u) N_j^q(v)w_{rj})^2}}} \Big({{\sum_{j=0}^m{N_j^q(v)w_{ij}}}}\Big)
	\label{eq:NURBSvDerivative6}   
\end{align}

Now, we evaluate the derivative of $N_i^p(u)$ with respect to the knot points $\{u_i\}$. We observe that due to the recursive nature of the definition, we can accordingly compute the derivatives of $N_i^p(u)$ in a recursive fashion using chain rule, \emph{provided} we can evaluate:
\begin{align}
\frac{\partial N_i^0(u)}{\partial u_i} = \frac{\partial \mathbf{1}([u_i,u_{i+1}])}{\partial u_i}
\end{align}
(and likewise for $u_{i+1}$) where $\mathbf{1}$ denotes the indicator function over an interval. However, this derivative is not well-defined since the gradient is zero everywhere and undefined at the interval edges.

We propose to approximate this derivative using \emph{Gaussian smoothing} by rewriting the interval as the difference between step functions convolved with deltas shifted by $u_i$ and $u_{i+1}$ respectively:
\begin{align}
\mathbf{1}([u_i,u_{i+1}))(u) = \text{sign}(u) \star \delta(u - u_{i}) - \text{sign}(u) \star \delta(u - u_{i+1})
\end{align}
and approximate the delta function with a Gaussian of sufficiently small (but constant) bandwidth:
\begin{align}
\mathbf{1}([u_i,u_{i+1}])(u) = \text{sign}(u) \star G_\sigma(u - u_{i}) - \text{sign}(u) \star G_\sigma(u - u_{i+1})
\end{align}
where
\begin{align}
G_\sigma(u - \mu) = \frac{1}{\sqrt{2\pi \sigma^2}} \exp(- \frac{(u-\mu)^2}{2\sigma^2})
\end{align}

The derivative with respect to $\mu$ is therefore given by:
\begin{align}
G_\sigma'(u=\mu) = \frac{(u - \mu)}{2\sigma^2} G_\sigma(u - \mu)
\end{align}
which means that the approximate gradient introduces a multiplicative $(u - \mu)$ factor with the original basis function. 

Therefore, now we can compute $\nicefrac{\partial N_i^p(u)}{\partial u_i}$ by recursively defining the derivatives $\nicefrac{\partial N_i^{0}(u)}{\partial u_i}, \nicefrac{\partial N_i^{1}(u)}{\partial u_i}$, until $\nicefrac{\partial N_i^{p-1}(u)}{\partial u_i}$. The derivative of $\nicefrac{\partial N_i^p (u)}{\partial N_i^{p-1} (u)}$ can easily be obtained from chain rule of  \eqnref{eq:NURBSBasis}.  Now, we perform the same operations of $\nicefrac{\partial N^p_i(u)}{\partial u_i} \forall u_i \in \vec{U}$ to obtain $\nicefrac{\partial N^p_i (u)}{\partial \vec{U}}$ and finally obtain $\nicefrac{\partial \vec{S}(u,v)}{\partial \vec{U}}$. Same operations can be performed to obtain $\nicefrac{\partial \vec{S}(u,v)}{\partial \vec{V}}$. With all these operations for each point parameterized in the surface by $(u,v)$, we extend this to all the surfaces as explained in the next section.

\subsubsection{Jacobian for Surface Evaluation}
We define the Jacobian for the NURBS evaluation, which is then directly used for the backward evaluation. The Jacobian for each surface evaluation point $S_{i,j}$ is represented as the vector:

\begin{equation} \vec{B}_{i,j} = \begin{pmatrix}
\{\vec{S}_{,p_{ij}}(u,v)\} \forall i \in [1,m], \forall j \in [1,n] \\
\{\vec{S}_{,w_{ij}}(u,v)\} \forall i \in [1,m], \forall j \in [1,n] \\
\{ \vec{S}_{,u_{i}}(u,v)\} \forall u_{i}, \in \vec{U}  \\
\{ \vec{S}_{,v_{j}}(u,v)\} \forall v_{j}, \in \vec{V}  \\
\end{pmatrix}\end{equation}

Each Jacobian vector represented here is the contribution of the gradient from one evaluation point at the grid locations $(i,j)$ in the parametric coordinate space. These Jacobian vectors are each of length $4nm$. However, as noted in the previous section on forward evaluation, the basis functions satisfy the partition of unity and span only $p+1$ control points starting from $u_{span}$ (correspondingly, $q+1$ control points starting from $v_{span}$ in the other parametric direction). Therefore, the total number of non-zero elements in a $4nm$ size Jacobian vector is $4(p+1)(q+1)$, making it sparse. However, note that this Jacobian is for only one surface point. The complete Jacobian for the backward pass is given as:
\begin{equation}
    J = \begin{pmatrix}
    \vec{B}_{1,1} \\
    \vec{B}_{1,2} \\
    \vdots\\
    \vec{B}_{m_{grid},n_{grid}}
    \end{pmatrix}.
    \label{eq:NURBSJacobian}
\end{equation}

The size of this Jacobian is $n_{grid}m_{grid}\times 4nm$. Here, $\vec{B}_{i,j}$ is the Jacobian for one surface point evaluation. As the parametric coordinates keep changing, the position of $u_{span}$ and $v_{span}$ keep changing, and the location of the non-zero elements keeps shifting to form a block diagonal matrix. This Jacobian is $\nicefrac{\partial{\vec{S}}}{\partial{\Psi}}$. For completing the backward pass, we multiply $\nicefrac{\partial{\mathcal{L}}}{\partial{\vec{S}}}$ to $\nicefrac{\partial{\vec{S}}}{\partial{\Psi}}$, giving us $\nicefrac{\partial{\mathcal{L}}}{\partial{\Psi}}$. Since, each module in the deep learning framework is independent and modular, we just return this output for the NURBS backward evaluation.

\subsubsection{Implementation}
For the implementation of the backward pass, since the basis functions are block sparse, we make use of the stored information of $u_{span}$ and $v_{span}$ for identifying the index of the control points derivative and we use the stored basis functions information for computing the Jacobian explained above. This computation is performed for all the surfaces in the batch. This complete algorithm is explained in detail in \algref{Alg:Backward}.

\begin{algorithm}[h]
    \caption{Backward Algorithm\label{Alg:Backward}}
    \SetKwInOut{Input}{Input}
    \SetKwInOut{Output}{Output}

    \Input{$\vec{S'}$}
    \Output{$\vec{P'}$, $\vec{W'}$}
    
    \text{Initialize: $\vec{P'} \rightarrow 0$}\\
    \text{Initialize: $\vec{W'} \rightarrow 0$}\\
    \For{$ k = 1: surfaces$}
    {
    \For{$j=1:m_{grid}$}
      {
        \For{$i=1:n_{grid}$}
          {
            Retrieve $u_{span}$, $v_{span}$, $N_{i}^p$, $N_{j}^q$, $\vec{S}(u,v)$\\
            \For{$r=0:p+1$}{
            \For{$h=0:q+1$}{
             $\vec{P'}_{u_{span}+r,v_{span}+h} = \vec{S}_{,p_{ij}}(u_i,v_j)$\\
             $\vec{W'}_{u_{span}+r,v_{span}+h} = \vec{S}_{,w_{ij}}(u_i,v_j)$\\
             $\vec{U'}_{u_{span}+r} = \vec{S}_{,u_{span}+r}(u_i,v_j)$\\
             $\vec{V'}_{v_{span}+h} = \vec{S}_{,v_{span}+h}(u_i,v_j)$\\
            }
            }
          }
      }
    }
\end{algorithm}

\subsection{GPU Implementation}\label{SubSec:GPUImplmentation}
We implemented the code in \emph{Python 3.6}~\citep{python36}. The backend for the GPU-accelerated code is written in \emph{C++} using the Pybind11 API~\citep{jakob2017pybind11} and CUDA toolkit~\citep{CUDA} for GPU acceleration and is integrated with PyTorch~\citep{paszke2019pytorch} using a custom layer definition. The forward evaluation can be performed for each surface in the batch for each tuple $(u,v)$ in the mesh grid of $u_{grid}\times v_{grid}$ in parallel. Further, the three coordinates $x$, $y$, $z$ are evaluated simultaneously. This enables an embarrassingly parallel implementation on the GPU for the forward evaluation of the \nd{} module. Each $x$, $y$, and $z$ component is mapped to a separate thread on the GPU  using the 3D block and grid structure in CUDA. The same process is employed in the backward algorithm with one additional operation. Each surface point gradient needs to be added to several control points that lie in the evaluated point's span during the backward pass. Hence we perform this operation of the gradient update using a \texttt{scatter} operation by using the indices stored from $u_{span}$ and $v_{span}$.

\section{CAD Applications using the \nd{} Module}\label{Sec:CADApplications}
The differentiable programming approach explained above is designed mainly for deep learning applications. However, we can also use the framework for standard CAD operations, such as curve fitting, surface fitting, and surface offsetting. Note that some of these operations could be performed much faster using traditional approaches explicitly optimized for each application. However, using \nd{} along with gradient-descent-based optimization approaches for these CAD applications is not well explored. Moreover, this shows the versatility of the \nd{} module in handling traditional CAD operations as constraints in a deep learning system. 

To use our differentiable programming approach for CAD applications, we have to define two key elements: a loss function $\mathcal{L}$ for computing the gradients and an optimization algorithm. We consider four loss functions: $\mathcal{L}_1$ loss, $\mathcal{L}_2$ loss (also called as mean squared error), the Chamfer distance $\mathcal{L}_{CD}$, and the Hausdorff distance $\mathcal{L}_{HD}$.

The $\mathcal{L}_1$ loss can be mathematically defined as: 
\begin{equation}
    \mathcal{L}_1(\vec{P}, \vec{Q}) = \frac{1}{n_{points}} \Big( \sum_{(\vec{P_i},\vec{Q_i}) \in (\vec{P},\vec{Q})} ||\vec{P_i} - \vec{Q_i}||_1 \Big)
    \label{eq:L1Distance}
\end{equation}

Here, $||\vec{P_i} - \vec{Q_i}||_1$ refers to the $L_1$ norm of the difference between the two points $\vec{P_i}$ and $\vec{Q_i}$. Similarly, we can define $\mathcal{L}_2$ loss based on the $L_2$ norm.
\begin{equation}
    \mathcal{L}_2(\vec{P}, \vec{Q}) = \frac{1}{n_{points}} \Big( \sum_{(\vec{P_i},\vec{Q_i}) \in (\vec{P},\vec{Q})} ||\vec{P_i} - \vec{Q_i}||_2 \Big)
    \label{eq:L2Distance}
\end{equation}

While both the $\mathcal{L}_1$ and $\mathcal{L}_2$ loss functions are pairwise distance metrics, the Chamfer distance ($\mathcal{L}_{CD}$) and the Hausdorff distance ($\mathcal{L}_{HD}$) are global distance metrics between two sets of points as shown below:
\begin{equation}
    \mathcal{L}_{CD} = \sum_{\vec{P_i}\in\vec{P}}{\,\min_{\vec{Q_j}\in\vec{Q}}{{||\vec{P_i}-\vec{Q_j}}||_2}} + \sum_{\vec{Q_j}\in\vec{Q}}{\,\min_{\vec{P_i}\in\vec{P}}{||{\vec{P_i}-\vec{Q_j}}||_2}}
    \label{eq:ChamferDistance}
\end{equation}
\begin{equation}
    \mathcal{L}_{HD} = \max_{\vec{P_i}\in\vec{P}}\left({\,\min_{\vec{Q_j}\in\vec{Q}}{{||\vec{P_i}-\vec{Q_j}}||_2}}\right) + \max_{\vec{Q_j}\in\vec{Q}}\left({\,\min_{\vec{P_i}\in\vec{P}}{||{\vec{P_i}-\vec{Q_j}}||_2}}\right)
    \label{eq:HausdorffDistance}
\end{equation}

For each CAD application, a target point cloud $\vec{Q}$ is obtained directly from measurements (for the case of fitting from point clouds) or from analytical computations (for CAD operations such as surface offsetting). While our formulation works well when we initialize the control points and weights to random values (Gaussian distributed), we can also initialize it using a prior for faster convergence. After we initialize the control points $\vec{P}$ and weights $\vec{W}$, we initialize the knot vectors $\vec{U}$ and $\vec{V}$. While the knot vectors are fixed in most applications we demonstrate, we show one example where we could reparameterize the surface to obtain a better fit. We evaluate the surface using $\vec{P}$, $\vec{W}$, $\vec{U}$, and $\vec{V}$ and compute the loss between the evaluated surface $\vec{S}$ and $\vec{Q}$ using the appropriate loss function $\mathcal{L}$. We perform an update using gradient descent algorithms and their variants. We can write a simple update for the NURBS parameters as:
\begin{equation}
    \Psi = \Psi - \alpha \frac{\partial\mathcal{L}}{\partial\Psi}
    \label{eq:SGD}
\end{equation}

While our formulation can use a simple gradient descent algorithm, in our work, we use more sophisticated algorithms such as stochastic gradient descent (SGD), SGD with momentum, Adam~\citep{kingma2014adam}, and Adagrad~\citep{lydia2019adagrad}. Our experiments illustrated in the Appendix show that SGD with momentum and Adam perform well in all scenarios and have faster convergence. Now, we discuss the specific CAD applications using our \nd{} module.

\subsection{Curve Fitting}\label{SubSec:2DFitting}
We first demonstrate curve fitting using our \nd{} module. While the problem of fitting splines to point clouds using data-driven techniques has been extensively studied, we show curve fitting to validate our approach and provide insights (such as the convergence of the optimization, behavior of the fit with the variation in control points, evaluation points, etc.). In this validation case, we initialize a uniform knot vector $\vec{U}$ (which is kept fixed) and compute the non-zero basis functions $N(u)$ and $u_{span}$ as a pre-computation step before evaluating the points on the curve. This pre-computation step reduces the computational time significantly (due to the removal of repeated basis function evaluations). This approach can be used when we are trying to fit a B-spline curve/surface where the basis functions do not change or if the user does not want to change the parameterization of the curve or surface. 

\begin{figure*}[b!]
    \centering
    \begin{subfigure}[b]{.3\linewidth}
        \centering
        \includegraphics[width=\textwidth,trim={0.0in 0.2in 0.0in 0.3in},clip]{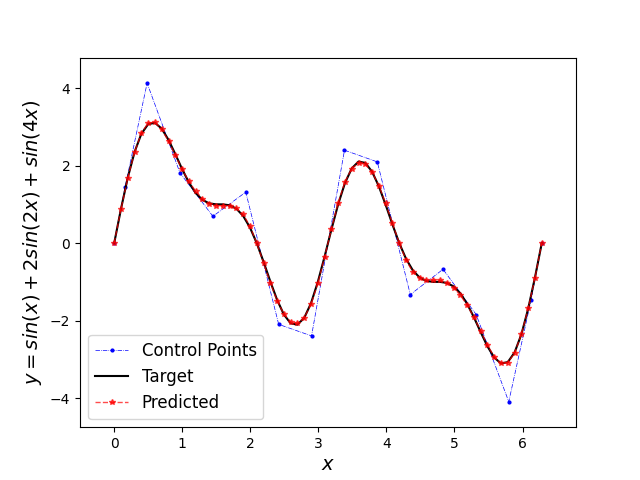}
        \label{Fig:16x64}
        \vspace{-0.2in}
        \caption{16 control points, 64 evaluation points}
    \end{subfigure}
    \begin{subfigure}[b]{.3\textwidth}
        \centering
        \includegraphics[width=\textwidth,trim={0.0in 0.2in 0.0in 0.3in},clip]{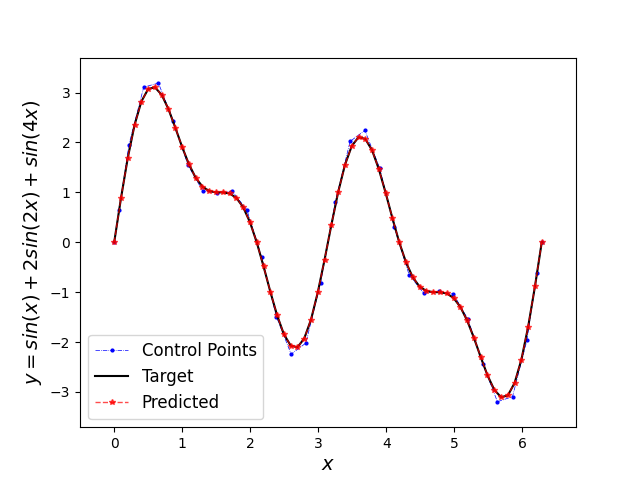}
        \label{Fig:32x64}
        \vspace{-0.2in}
        \caption{32 control points, 64 evaluation points}
    \end{subfigure}
    \begin{subfigure}[b]{.3\textwidth}
        \centering
        \includegraphics[width=\textwidth,trim={0.0in 0.2in 0.0in 0.3in},clip]{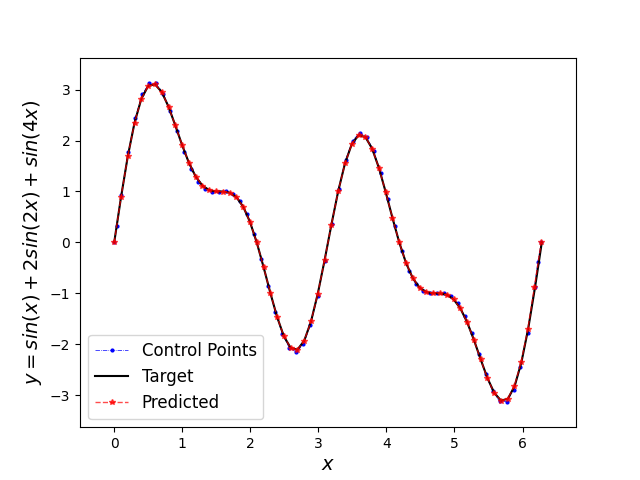}
        \label{Fig:64x64}
        \vspace{-0.2in}
        \caption{64 control points, 64 evaluation points}
    \end{subfigure}
    \caption{Curve fitting to points from an analytically generated curve $y=sin(x) + 2sin(2x) + sin(4x)$ with different number of control points.}
    \label{Fig:curve-fitting}
\end{figure*}

\begin{figure*}[b!]
    \centering
    \includegraphics[width=0.95\linewidth,clip,trim={0.0in 4.3in 0.0in 0.0in}]{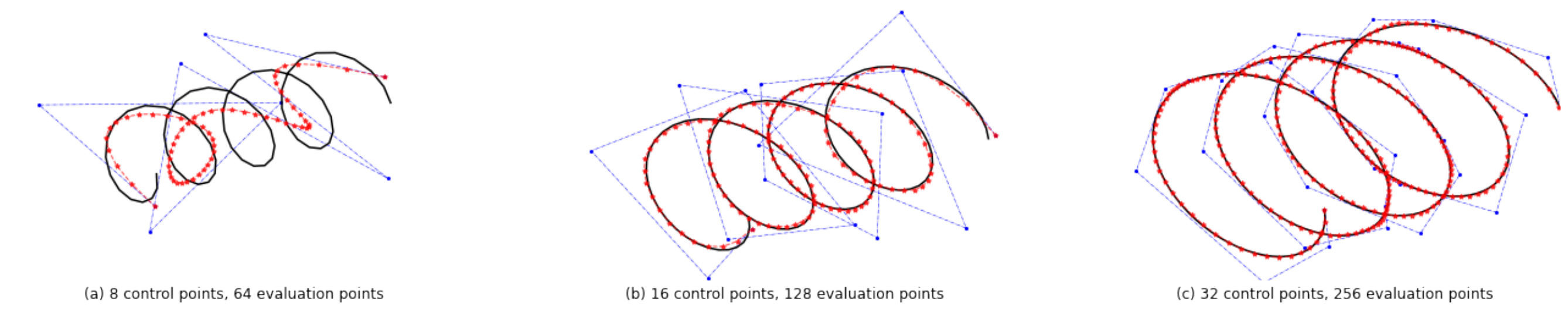}
    \includegraphics[width=0.95\linewidth,clip,trim={0.0in 0.05in 0.0in 7.3in}]{Figures/plotting_helix.pdf}
    \caption{Curve fitting for points sampled from a 3D helical curve in $\mathcal{R}^3$ for different numbers of control points and evaluated points.}
    \label{Fig:helix_fitting}
\end{figure*}

\begin{figure*}[b!]
    \centering
    \includegraphics[width=0.9\linewidth]{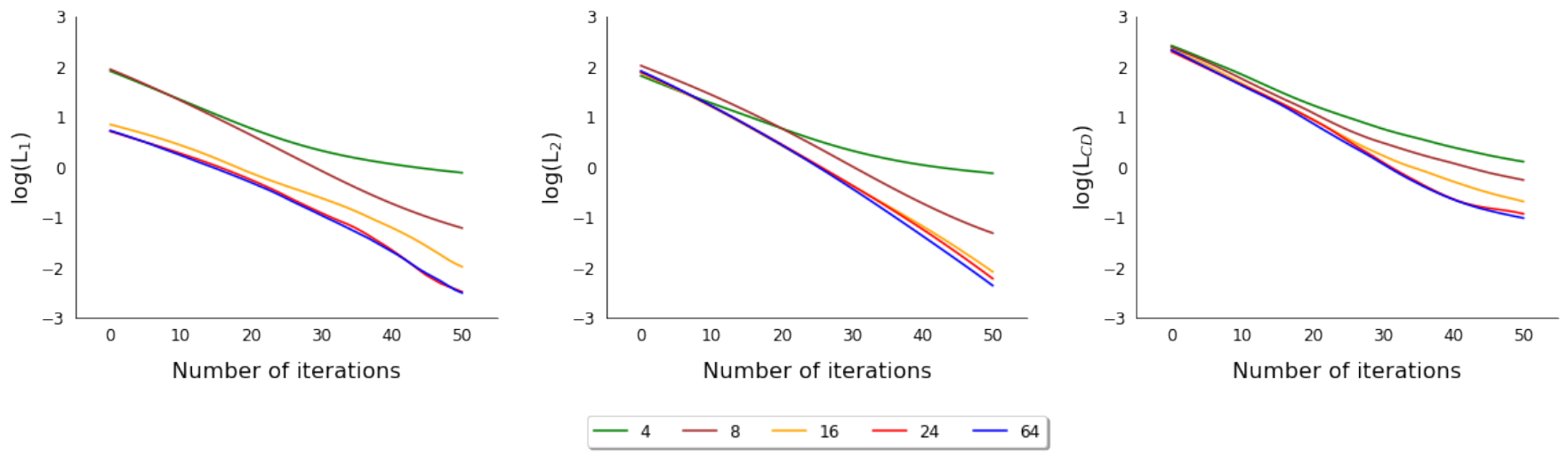}
    \caption{The loss performance with the number of iterations for fitting the 3D helical curve shown in \figref{Fig:helix_fitting}. The behavior of the loss is shown for different loss functions and different numbers of control points used for the fitting.}
    \label{Fig:helix_fitting_loss}
\end{figure*}

For a simple validation, we sample points from an analytically defined curve $y=sin(x) + 2sin(2x) + sin(4x)$ and obtain the best-fit curves for different control points and evaluation points as shown in \figref{Fig:curve-fitting}. Similarly, for points sampled from a 3D helical curve, we fit a B-Spline curve as shown in \figref{Fig:helix_fitting}. The behavior of our fitting method with different loss functions and the number of control points is shown in \figref{Fig:helix_fitting_loss}. While all the loss functions converge well with the number of iterations, $\mathcal{L}_2$ loss and $\mathcal{L}_{CD}$ losses have better convergence characteristics. $\mathcal{L}_{CD}$ loss plateaus after some iterations since that is the best error that could be achieved for a given number of control points and evaluation points. Please note that the $\mathcal{L}_{CD}$ also reduces the oscillations that might occur in fitting a higher degree curve.

\begin{figure*}[t!]
    \centering
    \includegraphics[width=0.9\linewidth]{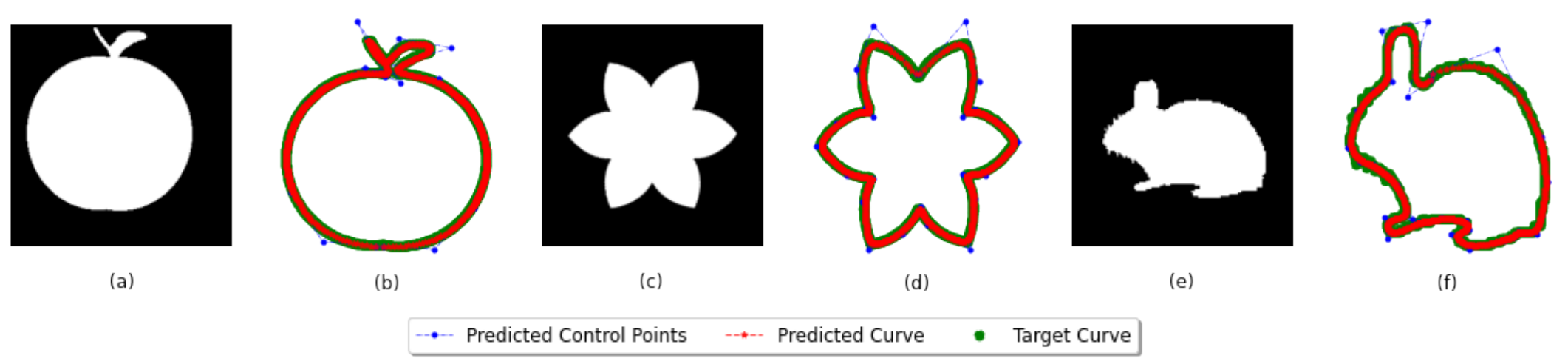}
    \caption{Curve fitting on point cloud data obtained from binary images. We convert the image data into a point cloud using the pixel size and the locations of the pixels. Once we obtain the point cloud, we use a curve-fitting module with Chamfer distance loss function and regularization to obtain these results.}
    \label{Fig:SkelnetonFitting}
\end{figure*}

\begin{table}[t!]
\centering
\caption{Performance ($\mathcal{L}_{CD}$) of fitting different curves using the \nd{} module with and without curve length regularization.}
\label{Tab:CurveFitting}
\newcommand{\tabincell}[2]{\begin{tabular}{@{}#1@{}}#2\end{tabular}}
\settowidth\tymin{Application}
\setlength\extrarowheight{2pt}
\begin{tabulary}{\linewidth}{|L|R|R|}
    \hline
    \textbf{Curve} & \textbf{No Regularization} & \textbf{Regularization}\\ \hline
    Analytical & 0.0025  &  \textbf{0.0016}   \\ 
    Helix & 0.0370  &\textbf{ 0.0320}  \\ 
    Apple & 8.5267  &  \textbf{1.6863} \\ 
    Flower & 23.2484 & \textbf{0.8383} \\ 
    Bunny &  50.0020 &  \textbf{1.4099}  \\ 
    \hline
    \end{tabulary}%
\end{table}

We extend the curve fitting framework to a more general fitting of random unordered point cloud data. For the 2D point cloud data, we use the images from the Pixel dataset of the Skelneton challenge~\citep{demir2019skelneton} and sample points from the object boundaries as shown in \figref{Fig:SkelnetonFitting}. Since the problem is ill-posed (due to the unordered aspect of the point cloud), we initialize the control points using randomly sampled points from the point cloud. To avoid unnecessary loops and self-intersections in our output curve, we use a curve-length regularization term in addition to the $\mathcal{L}_{CD}$. As shown in \figref{Fig:SkelnetonFitting} our framework fits the target well for point clouds that are not excessively complex. Some complex point cloud data that had self-intersections are illustrated in the Appendix.

We also analyze the performance of our fitting method for different curves in \tabref{Tab:CurveFitting}. For a consistent comparison, we use the $\mathcal{L}_{CD}$ between a dense set of points evaluated on the fitted curve (using 16 control points) and the input point cloud. We set the number of evaluation points to be twice the number of points in the input. Our \nd{} module achieves better fitting results with the curve length regularization added. While this difference is less evident for simple analytical curves, the advantage of using a curve length regularization is more pronounced over curves fitted using the Pixel dataset. 

\begin{figure*}[t!]
    \centering
    \includegraphics[width=0.85\linewidth, trim={0.0in 0.0in 0.0in 0.0in},clip]{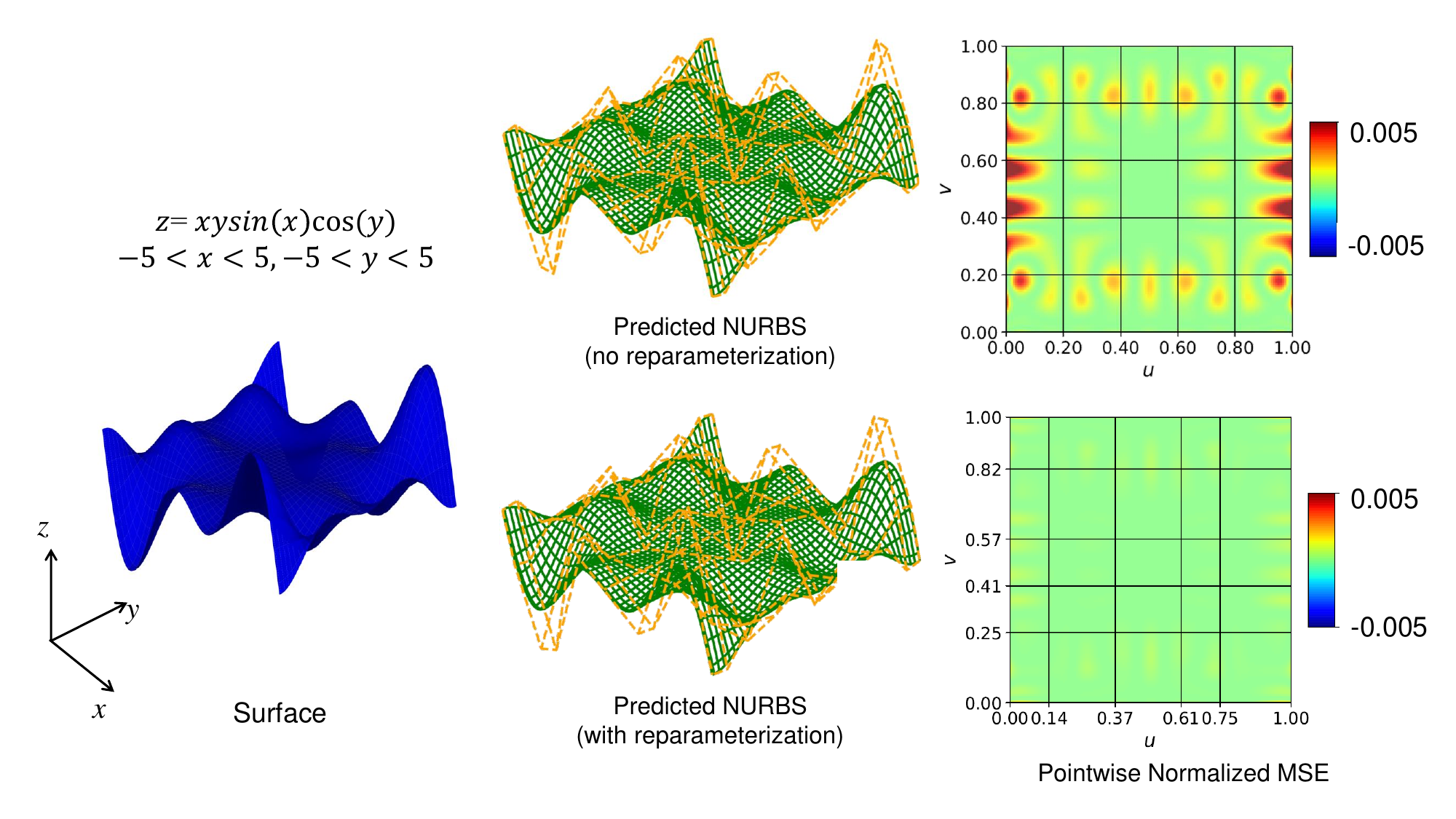}
    \caption{Surface fitting using NURBS for points sampled from an analytical surface $z = sin(x)*cos(y)$.}
    \label{Fig:surface_fitting}
\end{figure*}

\begin{figure*}[t!]
    \centering
    \includegraphics[width=0.8\linewidth,trim={0.5in 0.0in 0.3in 0.0in},clip]{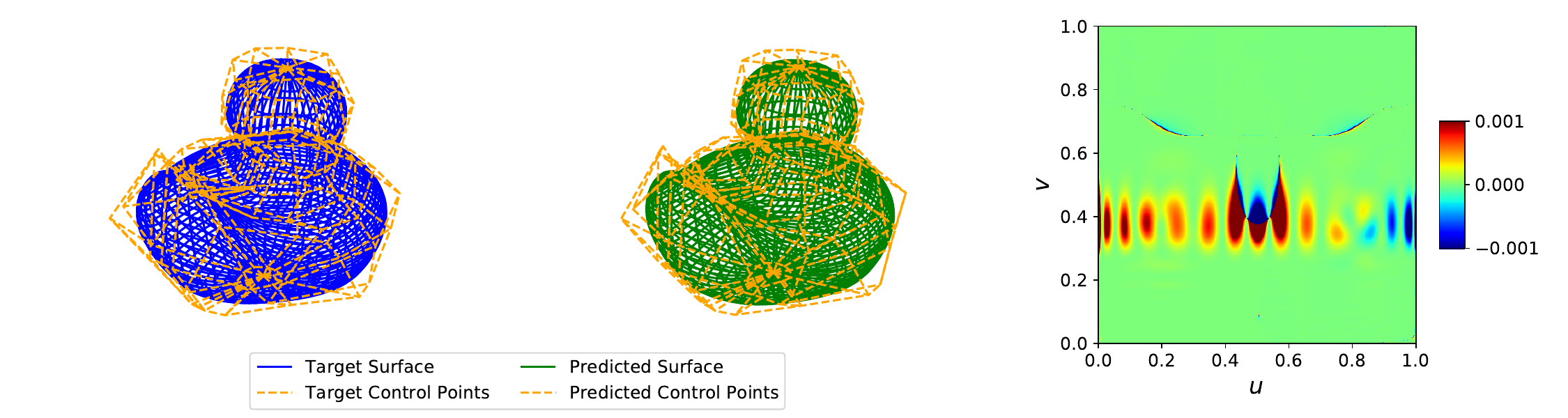}
    \caption{NURBS surface fitting for point cloud representation of Ducky's body.}
    \label{Fig:ducky_fitting}
\end{figure*}

\begin{table}[b!]
    \centering
    \caption{Performance $\mathcal{L}_2$ of the different surface CAD applications using the \nd{} module. B-Spline refers to fixed uniform knot vector, and NURBS refers to non-uniform knot vector obtained through optimization.}
    \label{Tab:SurfaceFitting}
    \newcommand{\tabincell}[2]{\begin{tabular}{@{}#1@{}}#2\end{tabular}}
    \settowidth\tymin{Application}
    \setlength\extrarowheight{2pt}
    \begin{tabular}{|l|c|c|}
        \hline
        \textbf{Test Case} & \textbf{B-Spline} & \textbf{NURBS} \\ 
        \hline
        Analytical  & 0.039528      & \textbf{0.005262}\\ 
        Ducky       &  0.000195     & \textbf{0.000180} \\
        \hline
    \end{tabular}
\end{table}

\subsection{Surface Fitting}\label{SubSec:3DFitting}

We extend the validation of our \nd{} module to the fitting of NURBS surfaces. In this case, we consider two scenarios for testing the fitting process. In the first scenario, we keep the knot vectors $\vec{U}$ and $\vec{V}$ fixed. In this scenario, we can precompute the basis functions $N_i^p(u)$ and $N_j^q(v)$ along with their respective span indices, $u_{span}$ and $v_{span}$). In the second scenario, we allow the knot vectors to be changed, leading to the reparameterization of the surface to obtain a better fit. We cannot precompute the basis functions in the second scenario since the knot vector is updated each iteration. We define a set of points sampled from the analytical surface $z =  x y sin(x)cos(y)$ to compare both scenarios. The control points and weights are initialized randomly for the fitting, while the knot vectors are initialized as uniformly spaced. Using the $\mathcal{L}_2$ loss for the optimization, we perform the fitting for both the scenarios as shown in \figref{Fig:surface_fitting}. While the first scenario provides a good fit, we can reduce the minor oscillation errors in the surface fit with the knot optimization to obtain an order of magnitude lower $\mathcal{L}_2$ loss for the second scenario, as shown in \tabref{Tab:SurfaceFitting}. Note that both the surface fits look similar and fit the target surface well, but the surface reparameterization reduces the overall error. The changes in the knot vector are depicted using the gridlines on the heatmap showing the error.

\begin{table}[t!]
\centering
\caption{$\mathcal{L}_2$ error for fitting different control point mesh sizes for the analytical surface with 128$\times$128 evaluation points and surface degree 3.}
\label{Tab:CtrlPtsMSE}
\newcommand{\tabincell}[2]{\begin{tabular}{@{}#1@{}}#2\end{tabular}}
\settowidth\tymin{Application}
\setlength\extrarowheight{2pt}
\begin{tabulary}{\linewidth}{|C|R|}
    \hline 
    \textbf{Control Points} & \textbf{$\mathcal{L}_2$ Error} \\ \hline
    {6$\times$6} & $2.2719 \times 10^{+1}$ \\ \hline
    {9$\times$9} & $3.9528 \times 10^{-2}$ \\ \hline
    {12$\times$12} & $6.9547 \times 10^{-4}$\\ \hline
    {24$\times$24} & $3.7953 \times 10^{-7}$\\ \hline
    {48$\times$48} & $1.0997 \times 10^{-7}$\\ \hline
\end{tabulary}
\end{table}

\begin{table}[t!]
\centering
\caption{$\mathcal{L}_2$ error for fitting a 9$\times$9 control mesh to the analytical surface for degree 3 with different numbers of evaluated points.}
\label{Tab:EvalPtsMSE}
\newcommand{\tabincell}[2]{\begin{tabular}{@{}#1@{}}#2\end{tabular}}
\settowidth\tymin{Application}
\setlength\extrarowheight{2pt}
\begin{tabulary}{\linewidth}{|C|R|}
    \hline 
    \textbf{Evaluation Points} & \textbf{$\mathcal{L}_2$ Error} \\ \hline
    {64$\times$64} & 0.0402 \\ \hline
    {128$\times$128} & 0.0395\\ \hline
    {256$\times$256} & 0.0391\\ \hline
    {512$\times$512} & 0.0389\\ \hline
\end{tabulary}
\end{table}

\begin{table}[t!]
\centering
\caption{$\mathcal{L}_2$ error for different degrees of a  9$\times$9 control mesh to the analytical surface using 128$\times$128 evaluation points.}
\label{Tab:DegreeMSE}
\newcommand{\tabincell}[2]{\begin{tabular}{@{}#1@{}}#2\end{tabular}}
\settowidth\tymin{Application}
\setlength\extrarowheight{2pt}
\begin{tabular}{|c|r|}
    \hline 
    \textbf{Degree} & \textbf{$\mathcal{L}_2$ Error} \\ \hline
    { 1 } & 0.3615 \\ \hline
    { 2 } & 0.0702  \\ \hline
    { 3 } & 0.0395 \\ \hline
    { 4 } & 0.0585 \\ \hline
\end{tabular}
\end{table}

We extend this surface fitting to a more complicated surface (the Ducky shown in \figref{Fig:ducky_fitting}). We have the target control points, weights, and knot vectors for this geometry. Using this information, we sample points from the geometry uniformly and then use that as a target for performing surface fitting. We only show the results for the best fit surface with knot vector optimization for brevity. We observe improvement in the fit by performing knot optimization as shown in \tabref{Tab:SurfaceFitting}. However, the improvement is not as pronounced as seen in the analytical geometry in this case.

Finally, we study the performance of our \nd{} module for fitting the analytical surface shown in \figref{Fig:surface_fitting} using a variety of control point sizes, evaluation point sizes, and degrees. To benchmark the performance of our module, we study the first 500 iterations of the optimization (both forward and backward pass) and report the $\mathcal{L}_2$ loss. The results of these experiments are highlighted under \tabref{Tab:CtrlPtsMSE}, \tabref{Tab:EvalPtsMSE} and \tabref{Tab:DegreeMSE}. We observe that 12 control points are required to fit the surface properly. As expected, increasing the number of control points reduces the $\mathcal{L}_2$ error. For the number of evaluation points, a similar trend is observed, where increasing the number of evaluation points reduces the $\mathcal{L}_2$ error. Similarly, the $\mathcal{L}_2$ error decreases as the degree of the surface is increased until it stabilizes, after which it shows a slight increase in the error probably due to overfitting oscillations. Particularly, in \tabref{Tab:DegreeMSE}, we notice that the error is higher with a lower degree due to being an underdetermined system. With increasing the degree, the fit improves, and then finally, at degree $4$, the system becomes overdetermined and hence overfits. 

\subsection{Surface Offsetting}\label{SubSec:Offset}
Generating an offset surface is one of the fundamental CAD operations. Traditionally an offset surface for NURBS is generated by first performing a B\`ezier decomposition of the NURBS surface and performing the offset for each patch. However, this changes the parameterization of the resulting offset surfaces. However, specific applications might require an offset surface with the same parameterization. We can easily use our \nd{} module to perform such an offset operation. Note that this approach only works for small offset distances such that the topology of the offset surface does not change. 

To perform the offset operation, we first compute a dense set of points and normals at specific $(u, v)$ and calculate the points on the corresponding offset surface by moving the points along the normal by the offset distance. We then fit a NURBS surface for the offset point cloud data using the same parameterization of the base surface using the surface fitting method (see \secref{SubSec:3DFitting}). 

Using our offsetting method, we can also generate offsets of objects consisting of multiple NURBS surfaces. Generating offset surfaces from multiple base surfaces requires applying constraints on the control points on the common edges of the base surfaces. While computing the surface normals, we identify the surface points along the shared edge. We then compute the average normals of these common points and then normalize them. Calculating the average normal allows us to ensure the continuity of the offset point cloud data. We then apply the constraints on the fitted control points of the common edges to be the same on both the surfaces that share the common edge. To ensure this continuity for the NURBS surfaces, we create a list of the shared control points. After the surface fitting iteration, the control points are updated; based on the shared list, i.e., the control points of the common edge are assigned the same coordinates. We perform this by computing the average of all the common points and setting them the same on both surfaces.

\begin{figure*}[t!]
    \centering
    \includegraphics[width=0.99\linewidth]{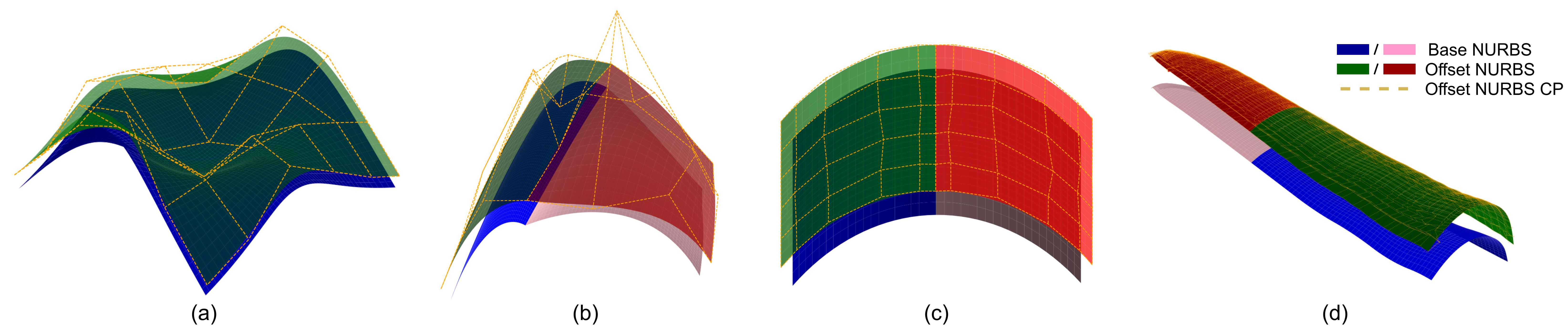}
    \caption{Test cases for NURBS surface offset (a) Double curved surface, (b) $C^0$ continuous multi patch, (c) $C^1$ continuous multi patch, and (d) $C^1$ continuous patches of the aerofoil profile of a wind turbine blade.}
    \label{Fig:OffsetTest}
\end{figure*}

\begin{table*}[t!]
\centering
\caption{Normalized Chamfer distance between the offset surface and the offset point cloud using our \nd{} module and SGD compared to a direct offset of the control points (CP Offset). The Chamfer distance is normalized using the minimum size of the bounding box of the base surface along the offset direction.}
\label{Tab:Offset}
\newcommand{\tabincell}[2]{\begin{tabular}{@{}#1@{}}#2\end{tabular}}
\settowidth\tymin{ Multi patch - $C^0$}
\setlength\extrarowheight{2pt}
\begin{tabulary}{\linewidth}{|L|R|R|R|R|}
    \hline
    \textbf{Test Case} & \textbf{Min BB Size} & \textbf{Offset Distance} & \textbf{$\mathcal{L}_{CD}$ \nd{} Offset} & \textbf{$\mathcal{L}_{CD}$ CP Offset} \\ \hline
    Double Curve        & 11.54     & 1.50      & \textbf{0.0235}      & 0.0239   \\ 
    Multi patch - $C^0$ &  0.75     & 0.10      & \textbf{0.0006}      & 0.0008   \\  
    Multi patch - $C^1$ &  6.32     & 2.00      & \textbf{0.0389}      & 0.0390   \\  
    Aerofoil Surface    &  0.66     & 0.25      & \textbf{0.0529}      & 0.0545   \\ \hline
    \end{tabulary}%
\end{table*}

\figref{Fig:OffsetTest} shows examples of the surface offsets generated using our \nd{} module. \subfigref{Fig:OffsetTest}{(a)} is a single surface patch with a double-curved surface. \subfigref{Fig:OffsetTest}{(b)} is a set of $C^0$ continuous surfaces with a single shared edge. \subfigref{Fig:OffsetTest}{(c)} is a set of $C^1$ continuous conic section surfaces. \subfigref{Fig:OffsetTest}{(d)} is a set of surfaces from a wind turbine blade model. To compare the accuracy of our \nd{} module approach for offset surfaces, we computed a simpler offset by offsetting the control points along the average normal direction of the control mesh. This approach works well for surfaces with low curvature. We then evaluated the computed offset surface at $25\times$ denser points than the number of points used for fitting and computed the $\mathcal{L}_{CD}$ with the input offset points. We find that our fitting approach achieves a lower $\mathcal{L}_{CD}$ than the control point offset approach for all cases, as seen in \tabref{Tab:Offset}.

\subsection{Timings for \nd{}}
In the previous sections, we demonstrate how the \nd{} module can perform CAD operations such as curve fitting, surface fitting, and surface offsetting. In this section, we assess the computational performance of our module. For brevity, we restrict our analysis to surface fitting operation and analyze the timings with variations in the number of control points, evaluation points, and surface degree. We only study the first 500 iterations (which include both the forward and backward pass). We perform all our experiments on a desktop with a 32 core 2.4 GHz Intel Xeon processor, 64 GB RAM, and an NVIDIA Titan Black GPU with 6 GB RAM.

\begin{table}[b!]
\centering
\caption{Time to fit a surface for different numbers of control points.}
\label{Tab:CtrlPtsTiming}
\newcommand{\tabincell}[2]{\begin{tabular}{@{}#1@{}}#2\end{tabular}}
\settowidth\tymin{Application}
\setlength\extrarowheight{2pt}
\begin{tabulary}{\linewidth}{|C|R|}
    \hline 
    \textbf{Control Points} & \textbf{Iteration time {(s)}} \\ \hline
    6 $\times$ 6  & 0.098 \\ \hline
    12 $\times$ 12  & 0.106\\ \hline
    24 $\times$ 24 & 0.110\\ \hline
    48 $\times$ 48 & 0.110\\ \hline
    % MSE  & 22719.6096 &	8.7109 & 	0.0728	& 0.0000     \\  \hline
\end{tabulary}%
\end{table}

\begin{table}[t!]
\centering
\caption{Time to fit a surface for different numbers of evaluation points.}
\label{Tab:EvalPtsTiming}
\newcommand{\tabincell}[2]{\begin{tabular}{@{}#1@{}}#2\end{tabular}}
\settowidth\tymin{Application}
\setlength\extrarowheight{2pt}
\begin{tabulary}{\linewidth}{|C|R|}
    \hline 
    \textbf{Evaluation Points} & \textbf{Iteration time {(s)}} \\ \hline
    64 $\times$ 64  & 0.074 \\ \hline
    128 $\times$ 128 & 0.120\\ \hline
    256 $\times$ 256  & 0.170\\ \hline
    512 $\times$ 512  & 0.266\\ \hline
\end{tabulary}%
\end{table}

\begin{table}[t!]
\centering
\caption{Computation time to fit a surface of different degrees.}
\label{Tab:DegreeTiming}
\newcommand{\tabincell}[2]{\begin{tabular}{@{}#1@{}}#2\end{tabular}}
\settowidth\tymin{Application}
\setlength\extrarowheight{2pt}
\begin{tabulary}{\linewidth}{|C|R|}
    \hline 
    \textbf{Degree} & \textbf{Iteration time {(s)}} \\ \hline
    1 & 0.074 \\ \hline
    2  & 0.120\\ \hline
    3  & 0.170\\ \hline
    4  & 0.266\\ \hline
    % MSE  & 22719.6096 &	8.7109 & 	0.0728	& 0.0000     \\  \hline
\end{tabulary}%
\end{table}

We analyze the timings against the different number of control points as shown in \tabref{Tab:CtrlPtsTiming}. We begin with the minimum number of control points we would get meaningful surfaces for the given experiment ( i.e., 6$\times$6 control points). We observe that the timings increase with the number of control points. Variations in iteration times for different evaluation point sizes are shown in \tabref{Tab:EvalPtsTiming}. Similar to \tabref{Tab:CtrlPtsTiming}, there is a steady increase in iteration times with an increase in the number of evaluation points. Note that, while the iteration time is increasing drastically (especially when going from $256\times256$ to $512\times512$), the performance of the fit does not improve much, as seen in \tabref{Tab:EvalPtsMSE}. Therefore, the end-user must judiciously choose the number of evaluation points sampled for the \nd{} module to get an accurate fit of the surface while not increasing the computational time. Finally, increasing the degree also increases the time required to fit a surface, as shown in \tabref{Tab:DegreeTiming}. However, for most cases, it can be seen the \nd{} module can perform 5-10 iterations per second, which makes it tractable for fitting a large number of surfaces.

\section{Point Cloud Reconstruction}\label{Sec:PCReconstructions}

In this section, we show the utility of the \nd{} module for unsupervised point cloud reconstruction. We use the experiments performed by \citet{sharma2020parsenet} as our baseline, which is a supervised learning framework for surface reconstruction framework. \citet{sharma2020parsenet} introduced an end-to-end trainable network called ParSeNet that fits an assembly of geometric primitives, including B-spline patches, to a segmented point cloud. In the ParSeNet framework, the authors develop a spline fitting module (called SplineNet) which takes an input point cloud and reconstructs a spline surface. This surface, however, is obtained using a supervised learning approach, as we highlight later. Therefore, to improve the framework and obtain a better fit without the supervised labels, we integrate our \nd{} module with SplineNet to develop an unsupervised training approach for spline fitting. 

The ParSeNet framework is divided into three stages. The first stage incorporates prior work done in point cloud segmentation \cite{qi2017pointnet} to decompose the input point cloud into segments classified under a parametric patch type. The second stage is the spline fitting SplineNet that generates  B-spline patches to the segmented point cloud data. The final stage performs geometric optimizations to seamlessly stitch the collection of predicted primitives together into a single object. We are interested in replacing the surface evaluation performed in the SplineNet stage of their network with our \nd{} module for the experiments in this section. For training and testing our experiments, we use the SplineDataset provided by \citet{sharma2020parsenet}. The SplineDataset is a diverse collection of open and closed splines that have been extracted from one million CAD geometries included in the ABC dataset. A random set of points is sampled from the surfaces as the input for the point cloud reconstruction task. We run our experiments on open splines split into 24K, 4K, and 4K for training, testing, and validation.

\begin{table*}[t!]
\centering
\scriptsize
\caption{Comparison between SplineNet~\citep{sharma2020parsenet} and our \nd{} implementation (with different number of control points). We compare the two-sided Chamfer distance (scaled by 100, 100$\times\mathcal{L}_{CD}$), the two-sided Hausdorff distance (scaled by 100, 100$\times\mathcal{L}_{HD}$)  between the input point cloud and a dense set of points sampled on the fitted surface, and also the Laplacian of the predicted control points $\mathcal{L}_{Lap}$ (scaled by 100, 100$\times\mathcal{L}_{Lap}$).}
\label{Tab:PCReconstruction}
\newcommand{\tabincell}[2]{\begin{tabular}{@{}#1@{}}#2\end{tabular}}
\setlength\extrarowheight{2pt}
\begin{tabulary}{0.99\linewidth}{|L|R|R|R|R|R|R|R|R|R|R|R|R|}
\hline
    \multirow{2}{*}{Loss Function} & \multicolumn{3}{c|}{\tabincell{c}{Baseline \\ ($20\times20$)} } &
    \multicolumn{3}{c|}{\tabincell{c}{ \nd{} \\ ($20\times20$) }} & 
    \multicolumn{3}{c|}{ \tabincell{c}{ \nd{} \\ ($5\times5$)} }  &
    \multicolumn{3}{c|}{\tabincell{c}{\nd{} \\ ($4\times4$)}} \\
    \cline{2-13}
    & $\mathcal{L}_{CD}$ & $\mathcal{L}_{HD}$ & $\mathcal{L}_{Lap}$ & $\mathcal{L}_{CD}$ & $\mathcal{L}_{HD}$ & $\mathcal{L}_{Lap}$ & $\mathcal{L}_{CD}$ & $\mathcal{L}_{HD}$ & $\mathcal{L}_{Lap}$ & $\mathcal{L}_{CD}$ & $\mathcal{L}_{HD}$ & $\mathcal{L}_{Lap}$ \\ \hline
    $\mathcal{L}_{CD} + \mathcal{L}_{LapMat} + \mathcal{L}_{CP}$& 1.184 & 0.994 & 1.340 &  &  &  &  &  &  &  & &   \\ \hline
    $\mathcal{L}_{CD} + 0.1 \mathcal{L}_{Lap} + 10 \mathcal{L}_{HD}$ &   &  &  & \textbf{0.018} & \textbf{0.234} & \textbf{0.202} & \textbf{0.020} & \textbf{0.256} & \textbf{2.462} & \textbf{0.028} & \textbf{0.274} & \textbf{3.356} \\ \hline
    $\mathcal{L}_{CD} + 0.1 \mathcal{L}_{Lap} + \mathcal{L}_{HD}$ &   &   & & 0.027 & 0.725 & 0.106 & 0.032 & 0.805 & 1.972 & 0.047 & 1.173 & 3.606  \\ \hline
    $\mathcal{L}_{CD} + 0.1 \mathcal{L}_{Lap}$ &  &  &  & 0.098  & 2.914 & 0.189  & 0.105  & 3.511  & 1.296 & 0.113 & 3.602 & 2.637   \\ \hline
    $\mathcal{L}_{CD}$ &   &   &   & 0.012 & 0.329 & 42.288 & 0.013 & 0.446 & 21.236 & 0.015 & 0.499 & 28.212  \\ \hline
    $\mathcal{L}_{HD}$ &   &   &   & 0.018 & 0.192 & 30.877 & 0.018 & 0.210 & 23.076 & 0.026 & 0.282 & 32.443  \\ \hline
\end{tabulary}%
\end{table*}

\begin{figure*}[b!]
    \centering
    \includegraphics[width=0.99\linewidth,trim={0.0in 1.9in 0.0in 1.9in},clip]{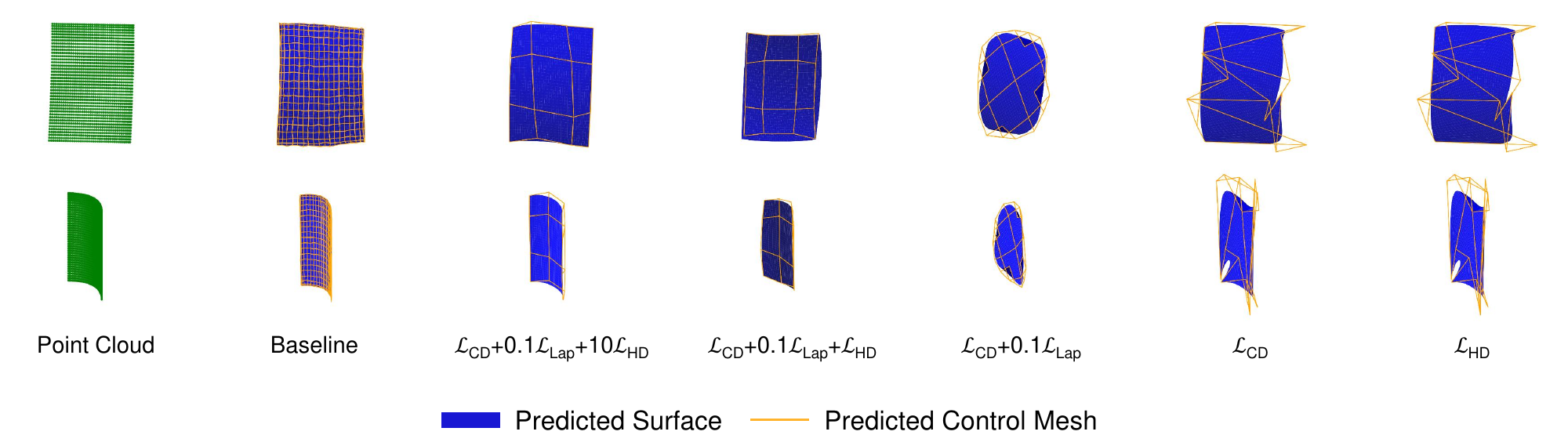}
    \caption{Comparison of the reconstructed B-spline surfaces obtained using \nd{} module for different loss functions using a 5$\times$5 control mesh.}
    \label{Fig:SurfaceReconstructionLossComparison}
\end{figure*}

In our work, we focus only on the SplineNet module of the ParSeNet framework to illustrate the applicability of our proposed \nd{} module. SplineNet is a module for performing supervised learning of B-spline surfaces using the SplineDataset. The training procedure includes a loss definition as follows:
\begin{equation}
    \mathcal{L}_{SplineNet} = \mathcal{L}_{CD} + \lambda_1 \mathcal{L}_{LapMat} + \lambda_2 \mathcal{L}_{CP}.
\end{equation}
Here, $\mathcal{L}_{CD}$ refers to the Chamfer distance between the input point cloud and the surface points evaluated on the target surface. $\mathcal{L}_{LapMat}$ is the Laplacian matching loss where the difference in the Laplacian (the second derivative of the control points mesh obtained using a Sobel filter) of the fitted control points and the target control points is minimized. $\mathcal{L}_{CP}$ is the control point regression loss which is $\mathcal{L}_2$ error between the predicted and the target control points. $\lambda_1$ and $\lambda_2$ are used for weighting different loss functions. For brevity, we use the same values for these constants as reported in \citet{sharma2020parsenet}.

\begin{figure*}[t!]
    \centering
    \includegraphics[width=0.98\linewidth,trim={0.0in 1.2in 0.0in 1.2in},clip]{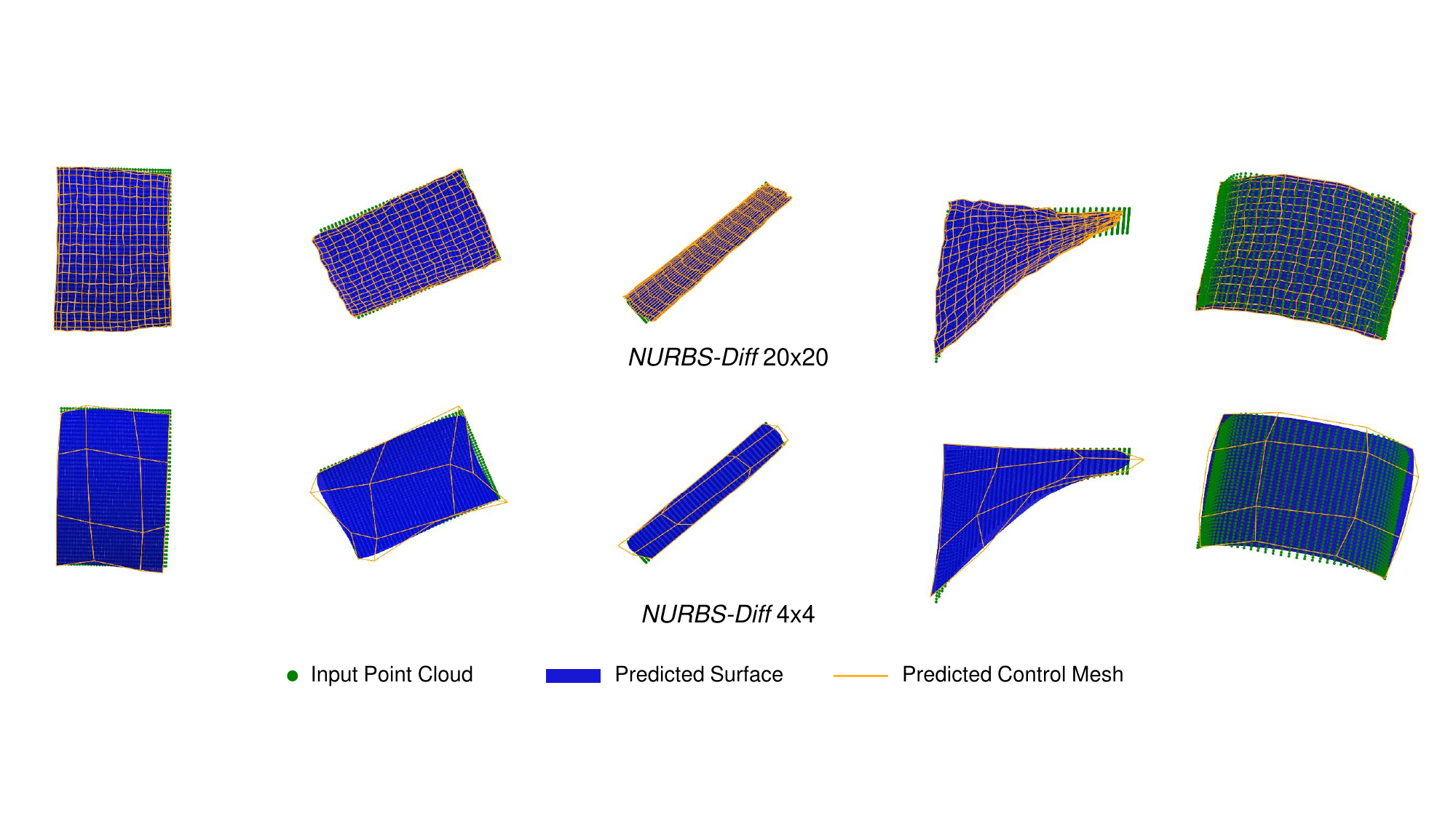}
    \caption{Point clouds and unsupervised reconstruction of B-spline surfaces (20$\times$20 and 5$\times$5) predicted by SplineNet using the \nd{} module.}
    \label{Fig:SurfaceReconstruction}
\end{figure*}

In the SplineNet approach, the $\mathcal{L}_{LapMat}$ and the $\mathcal{L}_{CP}$ make the framework supervised since they require a target fitted control point mesh to compute these losses. In real life, obtaining the target control point mesh for point clouds is challenging. Therefore, we need an unsupervised learning approach for point cloud reconstruction of spline surfaces. To make the framework unsupervised and not require the target control points of the spline surface, we modify the framework to the following:
\begin{equation}
    \mathcal{L}_{\nd{}} =  \mathcal{L}_{CD} + \lambda_1 \mathcal{L}_{Lap} + \lambda_2 \mathcal{L}_{HD}.
\end{equation}
Here, $\mathcal{L}_{CD}$ and $\mathcal{L}_{HD}$ refers to the Chamfer distance and the Hausdorff distance between the input point cloud and the surface points evaluated using the \nd{} module. $\mathcal{L}_{Lap}$ is a modified loss of $\mathcal{L}_{LapMat}$ where instead of matching the Laplacian of the predicted control points and actual control points, we minimize the Laplacian itself. Further, we scale down the contribution of $\mathcal{L}_{Lap}$ by an order of magnitude to reduce any adverse effects from minimizing the Laplacian. In addition, to fit the edges of the surfaces well, we add a Hausdorff distance loss to the objective function. We use different scale factors $\lambda_1$, $\lambda_2$ for $\mathcal{L}_{Lap}$ and $\mathcal{L}_{HD}$ to tune the objective function to obtain the best possible results.

Apart from the loss function defined above, the baseline supervised approach is restricted to non-rational B-spline surfaces because of no target weights. Here, in our case, we can perform the fitting for rational B-Spline surfaces or even B-Spline surfaces with reparameterized knots. Since the dataset available is very primitive and does not contain many complex structures, we restrict our analysis to just studying rational B-Splines reconstruction from the point clouds in an unsupervised manner.

\tabref{Tab:PCReconstruction} illustrates all the experiments we have performed for choosing an appropriate loss function. Performing surface reconstruction under no supervision is very challenging, and therefore several metrics have to be compared together to understand the performance. In \figref{Fig:SurfaceReconstructionLossComparison}, we illustrate how $\mathcal{L}_{CD}$ alone provides a bad fit by not covering the edges of the surfaces. Similarly, $\mathcal{L}_{HD}$ alone does not perform well. Therefore, we need a loss function that performs well in combination. We compare the results obtained from our approach with the baseline results obtained from \citet{sharma2020parsenet}, as illustrated in \tabref{Tab:PCReconstruction}. We observe that our approach performs an order of magnitude better than the baseline architecture in terms of $\mathcal{L}_{CD}$. Further, unlike \tabref{Tab:CtrlPtsMSE} where we have a complex analytical surface, the SplineDataset includes a collection of simpler open and closed surfaces. Therefore, we observe that reducing the number of control points required for representing the surface does not affect the fitting error as demonstrated by the (5$\times$5) and (4$\times$4) columns in \tabref{Tab:PCReconstruction}. Also, we note that using just $\mathcal{L}_{CD}$ and $\mathcal{L}_{HD}$ is not recommended due to high laplacian loss. Further, the combination of $\mathcal{L}_{CD} + 0.1\mathcal{L}_{Lap} + 10\mathcal{L}_{HD}$ gives the best result in terms of combined loss. We can also verify the same visually from \figref{Fig:SurfaceReconstructionLossComparison}. We also visualize a few anecdotal predicted surfaces along with the input point cloud in \figref{Fig:SurfaceReconstruction}. We see that the surfaces in the SplineNet dataset are not complex enough to require a large 20$\times$20 control point mesh and can be easily fitted with a small 4$\times$4 control point mesh. 

\section{Geometric Constraints using \nd{}}\label{sec:dlfea}
One of the advantages of our \nd{} module is the ease of enforcing surface constraints in deep-learning applications. We showcase the utility of our module in enforcing constraints using the example of valve deformation analysis. Analyzing bioprosthetic heart valves is essential for obtaining diagnostic information such as estimating the remaining life, fatigue, and patient-specific design. Traditionally, analysis of deformation behavior is performed using finite element or isogeometric analysis. However, such analyses are often computationally intensive. Recently, \citet{balu2019deep} proposed a deep learning framework for performing finite element analysis (called DLFEA) using a NURBS-aware convolutional neural network. Each heart valve is represented using three NURBS surface patches, and isogeometric analysis is performed to obtain the deformations for each control point under constant pressure applied during the valve closure. Fixed boundary conditions are applied on the valve edges that will be sutured to the aorta. Their previous work generated a large dataset of input geometry, pressure, thickness, and corresponding target deformations of the control points. However, their work does not perform a NURBS surface reconstruction loss for obtaining the deformations, which are more physically meaningful. Further, they did not explore the application of the fixed boundary conditions; it was implicitly enforced by modifying the loss function.

\begin{figure}[t!]
    \centering
    \includegraphics[width=0.99\linewidth, trim={0.3in 1.8in 0.3in  1.8in},clip]{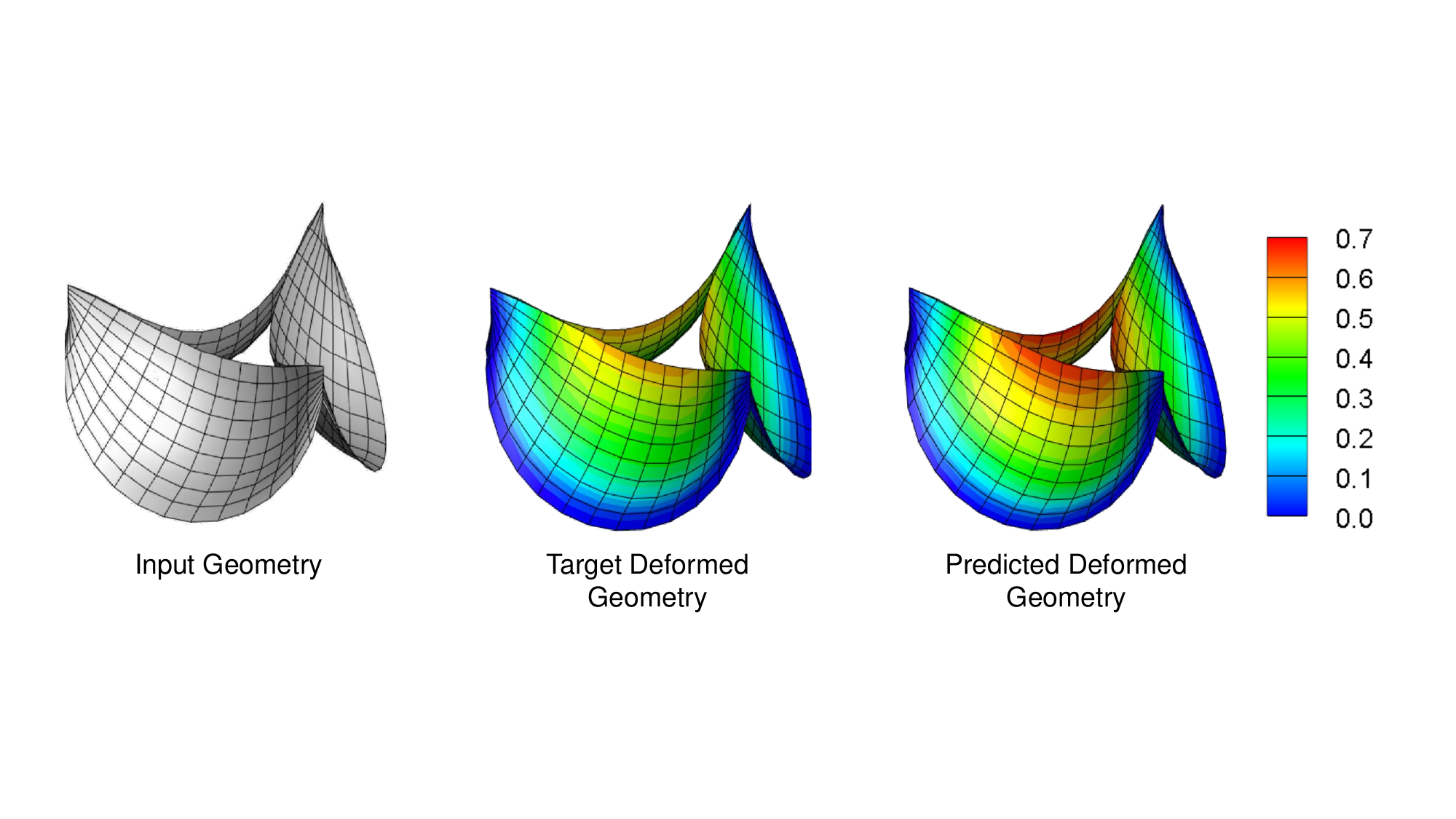}
    \caption{Visualization of an anecdotal bioprosthetic heart valve for DLFEA-BC from the test dataset. We show the input geometry, the target deformed geometry, and the predicted deformed geometry.}
    \label{Fig:DLFEA}
\end{figure}

In this work, we use the dataset and deep learning framework available from their work to demonstrate the utility of our \nd{} module in performing an evaluation of the deformations for the leaflets of the bioprosthetic heart valve. We perform a comparison between DLFEA, DLFEA with additional surface reconstruction loss (DLFEA-SR), and DLFEA with additional surface reconstruction loss and boundary condition enforcement loss (DLFEA-BC). We show the visualization of DLFEA-BC along with the input geometry, the target deformed geometry, and the predicted deformed geometry in \figref{Fig:DLFEA}. In DLFEA-BC, apart from the surface reconstruction loss, we add a constraint (BC loss) that the Dirichlet boundary conditions on the edge which is sutured to the aorta must be satisfied (i.e., the edges closest to the blue region in \figref{Fig:DLFEA} must have zero deformation).

In \tabref{Tab:DLFEA}, we show the results obtained on a test dataset (not used during the training). The CP loss is the $\mathcal{L}_2$ loss between the input control points and the target control points, whereas the surface reconstruction (SR) loss represents the $\mathcal{L}_2$ loss between the NURBS surface reconstruction of the target deformed shape and the actual deformed shape. We observe that DLFEA performs very well for CP loss (naturally because it was originally trained using that loss) but does not do well on the BC loss. The DLFEA-SR performs comparably for SR loss but has worse performance for BC loss. At the same time, DLFEA-BC performs the best for BC loss while performing comparably (although worse) on CP loss and SR loss. While the DLFEA-BC performs worse, the enforcement of boundary conditions makes it more physically meaningful. Further, in this ablation study, we show the result for each loss function applied independently. In general, we use a combined loss with several loss functions in tandem as done in \secref{Sec:PCReconstructions}.

\begin{table}[t!]
\centering
\caption{Performance comparison of DLFEA, DLFEA-SR, and DLFEA-BC. We compare the error on the test dataset for control points reconstruction, NURBS surface reconstruction, and boundary condition enforcement.}
\label{Tab:DLFEA}
\newcommand{\tabincell}[2]{\begin{tabular}{@{}#1@{}}#2\end{tabular}}
\settowidth\tymin{Application}
\setlength\extrarowheight{2pt}
\begin{tabular}{|l|r|r|r|}
    \hline
    \textbf{Test Case} & \textbf{CP loss} & \textbf{SR loss} & \textbf{BC loss} \\ \hline
    DLFEA & 7.79$\times 10^{-3}$       & 4.47$\times 10^{-3}$ & 37.69 \\ 
    DLFEA-SR & 12.84$\times 10^{-3}$    & 4.76$\times 10^{-3}$ & 46.14 \\
    DLFEA-BC & 12.83$\times 10^{-3}$    & 5.69$\times 10^{-3}$ & 10.90 \\
    \hline
\end{tabular}
\end{table}
% DLFEA & 0.007795 & 0.004472 & 37.692390 \\ 
% DLFEA-SR & 0.012845 & 0.004758 & 46.137878 \\
% DLFEA-BC & 0.012830 & 0.005686 & 10.901491\\

\section{Conclusions}\label{Sec:Conclusions}
We have developed a differentiable NURBS module that can be directly integrated with existing machine learning frameworks. We have developed a mathematical framework that enables both forward evaluation and backpropagation of the losses while training. Our module is GPU-accelerated to allow fast evaluation of NURBS surface points and fast backpropagation of the derivatives. We have demonstrated the utility of our NURBS module for several CAD applications and deep learning applications. \nd{} performs at the same level as existing standalone spline solutions used previously in the literature. Future work on the NURBS module includes developing support for trimmed NURBS surfaces and integrating complex curve constraints along trim edges for the watertight representation of CAD models. We have released the code for our NURBS module along with this paper. We believe this NURBS module will be the first step to better integrate deep learning with CAD and would lead to more diverse machine learning CAD applications.

\section*{Acknowledgement}
This work was partly supported by the NSF under grant CMMI-1644441 and the ARPA-E under DIF\-FERENTIATE:DE-AR0001215. This work used the Extreme Science and Engineering Discovery Environment (XSEDE), which is supported by NSF grant ACI-1548562 and the Bridges system supported by NSF grant ACI-1445606, at the Pittsburgh Supercomputing Center (PSC).

\pagebreak

\bibliographystyle{cag-num-names}
\bibliography{CAD2021}

\pagebreak

\appendix
\section{Parametric Derivatives}
Here we provide the parametric surface derivatives with respect to the parameters $u$ and $v$ for completeness. Please refer to \citet{10.5555/265261} for details.
\begin{equation}
	\vec{S}_{,u}(u,v)= \frac{\vec{NR}_{,u}(u,v)w(u,v) - \vec{NR}(u,v)w_{,u}(u,v)}{w(u,v)^2}
	\label{eq:NURBSuDerivative1}
\end{equation}
where,
\begin{equation*}
	\vec{NR}_{,u}(u,v)=\sum_{i=0}^n{\sum_{j=0}^m{N_{i,u}^p(u) N_j^q(v)w_{ij}\vec{P}_{ij} }}
\end{equation*}
\begin{equation*}
{w}_{,u}(u,v) = \sum_{i=0}^n\sum_{j=0}^m{N_{i,u}^p(u) N_j^q(v)w_{ij}}
\end{equation*}

\section{Test Surface Details}
\tabref{Tab:ModelDetail} shows geometrical details for each surface used for the fitting and surface offset tests. The surface fitting examples (Analytical and Ducky) are evaluated at $128^2$ and $512^2$ points, respectively. The surface offset test examples are evaluated at $20^2$ points each.

\begin{table}[h!]
\centering
\setlength\extrarowheight{3pt}
\caption{Test NURBS surface parameters.}
\label{Tab:NURBSModels}
\begin{tabular}{|l|r|r|r|r|}
\hline
\textbf{Surface Model} & \textbf{$p$} & \textbf{$q$} & \textbf{$n$} & \textbf{$m$} \\ \hline
Analytical & 3 & 3 & 12 & 12 \\
Ducky & 3 & 3 & 14 & 13 \\ 
Double Curve & 3 & 3 & 6 & 6 \\ 
Multi Patch - $C^0$ & 3 & 3 & 4 & 4 \\ 
Multi Patch - $C^1$ & 3 & 3 & 6 & 6 \\ 
Aerofoil & 3 & 3 & 50 & 24 \\ \hline
\end{tabular}
\label{Tab:ModelDetail}
\end{table}

\section{Computing Normals on the Common Edge}

\begin{figure*}[h!]
    \centering
    \includegraphics[width=0.6\linewidth]{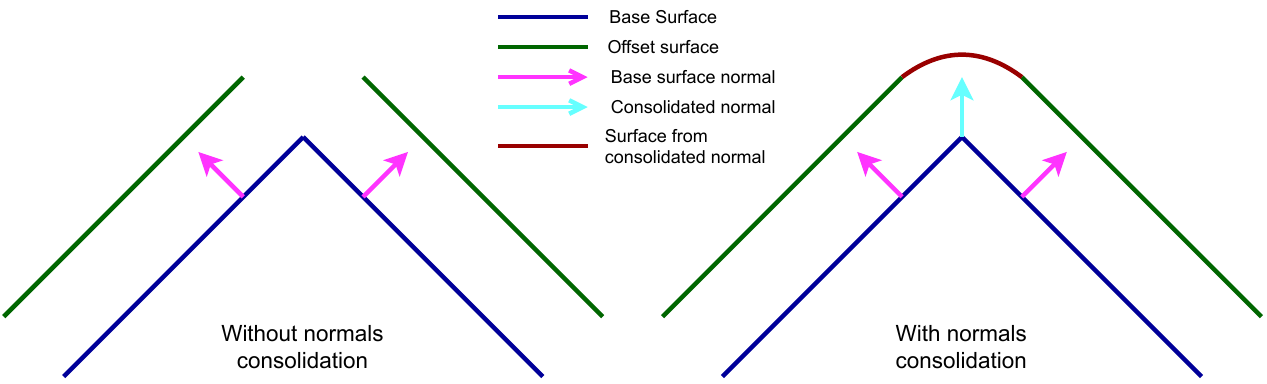}
    \caption{Without consolidated normals may lead to discontinuous surfaces, especially for $C^0$ connectivity (shown on the left). Normals consolidation ensures connectivity for the offset surfaces (shown on the right).}
    \label{Fig:NormalConsolidation}
\end{figure*}

For $C^0$ continuous surfaces, the surface normals at the common edge of two surfaces point in different directions. Hence, if the individual surface normals are used for the offset operation, it might lead to a gap or self-intersection between the offset surfaces. To deal with this case, we identify the common control points at the surface edges, and we recompute the resulting normal as the average of both the individual surface normals as shown in \figref{Fig:NormalConsolidation}. We then use this average normal to move the points on the common edges to generate the offset points.

\section{Additional Curve Fitting Results}

\begin{figure*}[t!]
    \centering
    \includegraphics[width=0.9\linewidth]{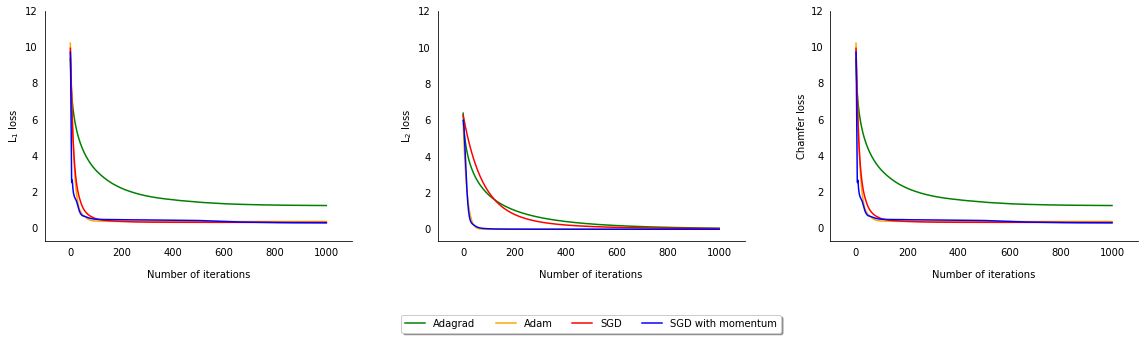}
    \caption{Results of the curve-fitting on a 3D helix point data with different optimizers.}
    \label{Fig:CurveFittingOptimizer}
\end{figure*}

We tested the curve fitting example of the 3D helix shown in \figref{Fig:helix_fitting} with different optimizers. \figref{Fig:CurveFittingOptimizer} shows the results for four different optimizers used in this test. We can see that all the optimizers converge to the same value for the $\mathcal{L}_2$ loss, but the Adagrad optimizer converges to a different value for the $\mathcal{L}_1$ and $\mathcal{L}_{CD}$ loss. In addition, SGD with momentum has the fastest convergence rates for all losses.

\begin{figure*}[t!]
    \centering
    \includegraphics[width=0.9\linewidth]{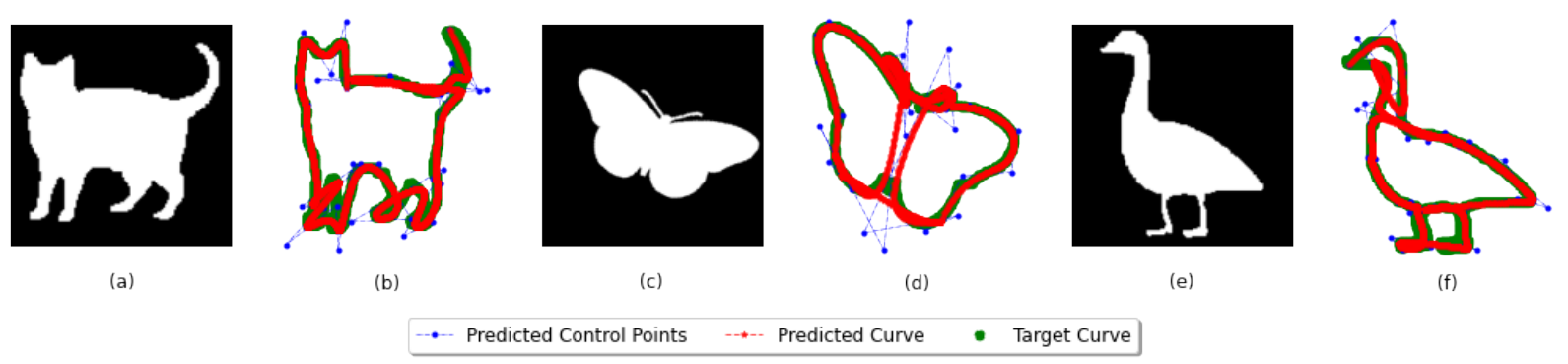}
    \caption{Curve fitting test performed on various skeleton geometry which results in an invalid curve generation due to self-intersecting geometry.}
    \label{Fig:SkelnetonLoops}
\end{figure*}

For some complex shapes, our method cannot generate a curve without self-intersections and loops for the Pixel dataset. This is because the weightage of the curve length regularization parameter needs to be tuned for each object based on its complexity. \figref{Fig:SkelnetonLoops} shows the results from 3 curve fitting tests where the curves generated are self-intersecting. Adding additional constraints to prevent this is a possible future research direction.

\section{\nd{} Implementation Details}

Our \nd{} module was implemented using \href{https://pytorch.org/}{Pytorch} library, which allows us to implement custom deep learning layers that we can use alongside traditional deep learning layers such as convolution, max-pooling, dense layers, etc. \texttt{torch.nn.module} gives us an interface to create our custom layers; however, the functions have to be defined in an automatically differentiable manner. In our application, we use \texttt{torch.autograd.Function} to define our custom forward and backward pass computations for the NURBS basis functions evaluation and curve/surface evaluation. This function is now used in the \texttt{torch.nn.module} to create the layer. The actual code for the forward and backward pass (for evaluation) is written in \texttt{C++} Language with \texttt{CUDA} integration for GPU acceleration. These modules (written in \texttt{C++}) are compiled along with \texttt{PyBind11} package to use in Python (supported by \href{https://pytorch.org/}{Pytorch}). The compilation is performed using a simple setup from \texttt{torch.utils.cpp\_extension}. The main source code of \nd{} module will be made public. 

We create two different layers: (i) B-Spline evaluation (ii) NURBS evaluation. Although the functions used in both are the same, in B-Spline evaluation, we avoid computing the basis functions every forward pass. Instead, we precompute the basis functions initially and only perform the forward evaluation of the curve/surface. This saves us computational time during the forward and backward pass. 

\end{document}